\documentclass[acmsmall,nonacm]{acmart} 
\AtBeginDocument{%
  }

\setcopyright{cc}
\setcctype{by}
\copyrightyear{2026}
\acmYear{2026}
\acmDOI{}

\acmConference[]{}{}{}

\acmBooktitle{}

\acmISBN{}

\usepackage{fontawesome}
\usepackage{pdflscape}
\usepackage{hyphenat}
\usepackage{enumitem}
\usepackage{spverbatim}
\usepackage{subcaption}
\usepackage{listings}
\usepackage{xcolor}
\usepackage{multirow}

\usepackage[table]{xcolor}

\definecolor{codegreen}{rgb}{0,0.6,0}
\definecolor{codegray}{rgb}{0.5,0.5,0.5}
\definecolor{codepurple}{rgb}{0.58,0,0.82}
\definecolor{backcolour}{rgb}{0.95,0.95,0.96}

\lstdefinestyle{mystyle}{
    backgroundcolor=\color{backcolour},   
    commentstyle=\color{codegreen},
    keywordstyle=\color{magenta},
    numberstyle=\tiny\color{codegray},
    stringstyle=\color{codepurple},
    basicstyle=\ttfamily\footnotesize,
    breakatwhitespace=false,         
    breaklines=true,                 
    captionpos=b,                    
    keepspaces=true,                 
    numbers=left,                    
    numbersep=5pt,                  
    showspaces=false,                
    showstringspaces=false,
    showtabs=false,                  
    tabsize=2
}

\begin{document}

\title{A Baseline Assessment for the Agentic Web Technology Stack}


\author{Vagner Figueredo de Santana}
\orcid{0000-0003-0325-1596}
\affiliation{%
  \institution{NIC.br}
  \city{São Paulo}
  \country{Brazil}}
\email{vsantana@nic.br}

\author{Diogo Cortiz}
\orcid{0000-0002-5875-8602}
\affiliation{%
  \institution{NIC.br / PUC-SP}
  \city{São Paulo}
  \country{Brazil}}
\email{diogo@nic.br}

\renewcommand{\shortauthors}{Santana et al.}

\begin{abstract}
The fast pace in which protocols, frameworks, and technologies are being proposed for LLM-based agents results in a complex scenario for anyone aiming to create state-of-the-art AI solutions.
Without an assessment on how different protocols, frameworks, and libraries work together to compose an Agentic Web solution, developers and organizations face different challenges (e.g., skilling, governance, infrastructure) even before becoming part of the Agentic Web ecosystem. 
In this context, we propose a system for performing Agentic Web evaluations called MASTIF (Multi-Agent System TestIng Framework) and present a baseline assessment of Agentic Web technologies, including 3 LLMs, 3 protocols, and 4 agent implementation frameworks, against 300 \textit{Mind2Web} benchmark tasks.
Results suggest that selecting a task-specific technology stack may result in gains of task completion rate, efficiency score, and social value stability scores; and reductions in verbosity and time to complete tasks. 
With the presented results, we expect to present a baseline in terms of metrics and results for this complex and fast evolving scenario for those aiming at designing, assessing, and developing multi-agent systems.

\end{abstract}




\maketitle

\section{Introduction}
\label{sec:introduction}

In classical AI literature, an agent is \textit{``anything that can be viewed as perceiving its environment through sensors and acting upon that environment through actuators''} \cite{Russell1995}.
Agents that can act beyond their built-in knowledge are told to be autonomous \cite{Russell1995}.
In an effort to differentiate between software program and agents, \citet{Franklin1996} proposed that `\textit{`An autonomous agent is a system situated within and a part of an environment that senses that environment and acts on it, over time, in pursuit of its own agenda and so as to effect what it senses in the future.''}
With recent advances in AI (Artificial Intelligence) and Large Language Models (LLMs), the goals envisioned with the idea of \textit{perceiving} and \textit{acting} were augmented by a natural language space \cite{Yao2022}. 
The capabilities brought by LLMs applied to AI agents powered the now popular LLM-based agents. These agents may act on model of their environments, learn about their operation, or involve complex orchestration \cite{Stryker2025}, being more useful for tasks that are harder, more realistic, and usually don't have a single correct answer \cite{Kapoor2024}.
In the content that follows, we use \textit{LLM-based agents} and \textit{agents} interchangeably to align with current research in the area.

With the emergence of LLMs, the information previously available only on static Web pages is now embedded in LLMs themselves \cite{Yang2025}.
In addition, approaches such as Chain-of-Thought (CoT) \cite{Wei2022} and \emph{ReAct} \cite{Yao2022} lead agents to perform intermediate reasoning steps, enabling planning and acting capabilities.
With this transformation in place, the Web is transitioning from the ecosystem where only (human) users consume and create content/services to the \textit{environment} in which agents are also part of this environment, \textit{perceiving} and \textit{acting} to perform high level, complex tasks, leading to the Agentic Web.

\newpage

\textbf{Agentic Web} can be defined as the ecosystem in which autonomous agents plan, coordinate, and execute goal-directed tasks, enabling continuous agent-agent interaction and value creation alongside human-Web interactions \cite{Yang2025}.
The term agentic was proposed as a way to cover the whole spectrum of use of agents rather than defining what an agent is or isn't \cite{Kapoor2024}.
Agent-agent communication here is considered to be the interaction that might not be rendered for human eyes, i.e., a new layer that involves optimized communication for agents in terms of parsing, planning, reasoning, and orchestration. Its core aspects include \cite{Yang2025}: 
\begin{enumerate}
    \item Agents must function autonomously performing complex tasks independently,
    \item Web resources need to be accessible to machines, and
    \item Value must be exchanged between humans, systems, and agents.
\end{enumerate}

Given that LLMs can use external APIs and tools in a self-supervised way \cite{Schick2023}, it is expected that the agent-agent interaction to increase, bringing challenges, opportunities, and risks, as this new layer is placed between the (human) end-user and the content/services available on the Web to be consumed/created by (digital) agents.
This agent-agent interaction can also be referred to as \textbf{agent-as-user}, in which agents operate autonomously as users of the Web, executing tasks, navigating interfaces, and completing workflows \cite{Yang2025}. Additional models of interaction include \textbf{agent-as-interface}, where agents enhance traditional user interfaces (e.g., browser), and \textbf{agent-with-physics}, where agents perceive, reason, and act in the physical world \cite{Yang2025}.

In many modern systems, autonomous agents do not operate alone but as \textbf{multi-agent systems} (MAS), coordinating, negotiating, or delegating tasks among a team of specialized agents. Some agents act as \textit{``orchestrators''} that manage workflows, assign sub-tasks, or monitor progress across sub-agents. These agents combine perception, planning, memory, and coordination faculties to drive large-scale or hierarchical task completion \cite{Stryker2025}. 
MAS have also proven value for open-ended research tasks and for breadth-first queries, pursuing multiple directions simultaneously \cite{Anthropic2025}.
Results suggest that MAS with 4 agents outperform by 90\% single agent for such tasks \cite{Anthropic2025}.
However, the use of agents has impacts in terms of token usage. 
In practice, single agents spent four times more tokens than LLM-powered chatbots and MAS spent up to fifteen times more tokens than LLM-powered chatbots \cite{Anthropic2025}. This also highlights a key challenge, i.e., economic viability for MAS, requiring high value tasks to justify its use for the increased performance \cite{Anthropic2025}.
In Agentic Web, MAS potentially involve agents interacting with APIs, browsers, or data services to accomplish tasks beyond  agent's internal reasoning.

Multiple protocols, frameworks, and tools are being proposed~\footnote{\url{https://github.com/kaushikb11/awesome-llm-agents}}$^{,}$\footnote{\url{https://www.promptingguide.ai/research/llm-agents}}. However, this fragmentation makes it hard for developers and organizations to onboard agentic systems or simply to keep up with technologies for agents that work together, effectively.
In this fast-paced complex evolving scenario \cite{Yang2025, Yang2025b}, this work aims to answer the following research questions (RQ) under the evaluated conditions for a baseline assessment: 


\begin{enumerate}[label=RQ\arabic*:]
    \item What is the most effective (task completion rate) Agentic Web technology stack?
    \item What is the most efficient (task completion rate, convergence time, and verbosity) Agentic Web technology stack?
    \item What is the Agentic Web technology stack that is best suited for tool-calling on the Web?
    \item What is the most stable Agentic Web technology stack in terms of social values embedded in their communication (e.g., prompts, reasoning, and output)?
\end{enumerate}

\newpage

In this context, this work contributes with: 
\begin{enumerate}
    \item A system for performing empirical assessment of Agentic Web technology stack called MASTIF (Multi-Agent System TestIng Framework);
    \item An empirical assessment baseline for the Agentic Web technology stack.
\end{enumerate}

This paper is structured as follows: 
section \ref{sec:related_work} reviews the state-of-the-art on Agentic Web, 
section \ref{sec:problem_definition} formulates the problem, metrics, and use case considered in this paper, 
section \ref{sec:experiment_setup} details the experiment performed, 
section \ref{sec:results} presents results obtained,
section \ref{sec:discussion} articulates results towards the research questions guiding this research, and
section \ref{sec:conclusion} concludes.

\section{Related Work}
\label{sec:related_work}

The recent wave of research on LLM agents has produced significant advances in reasoning, tool use, evaluation, and emerging interoperability standards. We group the most relevant works into the following main themes: types of agents, architecture, protocols, benchmarks, multi-agent system evaluations, and frameworks.

\subsection{Types of Agents}

Agents can be categorized along a spectrum of sophistication, memory, planning, and learning capabilities. The definitions presented next are useful for framing features and architectures discussed throughout this paper when assessing and comparing different technology stacks. 

\subsubsection{Simple reflex agents}  
These agents act purely on predefined condition–action rules, without memory or internal modeling, i.e., they map current percepts directly to actions. While suitable for very constrained or predictable domains, they cannot plan ahead or adapt based on previous experience \cite{Russell1995, Stryker2025}.

\subsubsection{Model-based reflex agents}  
These agents have an internal representation (model) of their environment and update it over time, allowing better reactivity than reflex-only systems by considering inferred state changes. However, they are still fundamentally reactive and lack explicit goal-driven planning or long-term foresight \cite{Russell1995, Stryker2025}.

\subsubsection{Goal-based agents}
These agents explicitly reason about future states to choose sequences of actions that achieve specified goals. Planning is central to their decision-making, which allows them to decompose tasks and make multi-step decisions toward achieving objectives—even when the path is not obvious from immediate observations \cite{Russell1995, Stryker2025}.
\textit{When no learning is involved, LLM-based agents often fall within this category.}

\subsubsection{Utility-based agents}  
Beyond simply reaching a goal, utility-based agents evaluate and compare potential outcomes using a utility function, seeking not just any solution but the \textit{most optimal} one according to specified criteria (e.g., speed, cost, accuracy, user satisfaction). This allows richer, more nuanced decision-making in tradeoff-laden environments \cite{Russell1995, Stryker2025}.

\subsubsection{Learning agents}  
Learning agents refine their behaviors over time through feedback, experience, or training. They can utilize supervised, reinforcement, or unsupervised learning to update their internal models, policies, or plans—thereby improving performance in dynamic, uncertain, or previously unseen scenarios. 
\textit{With learning in place, this category overlaps with LLM-based agents or other adaptive models \cite{Stryker2025}.}

\newpage

\subsection{Architecture}

\subsubsection{AGNTCY}
AGNTCY is an open source project aiming at structuring the Internet of Agents (IoA) by recommending a framework including\footnote{\url{https://agntcy.org/}}:
(1) agent discovery, using Open Agent Schema Framework (OASF) to allow agents to discover and understand the capabilities of others.
(2) agent identity, providing verifiable identity and access control across organizational boundaries, 
(3) agent messaging, supporting communications via SLIM (Secure Low-latency Interactive Messaging), 
(4) agent observability, providing end-to-end provenance in MAS,
(5) protocol integration, supporting agent and model protocols.

\subsubsection{Networked AI Agents in Decentralized Architecture (NANDA)}
Project NANDA proposes an infrastructure for an open Agentic Web by supporting how AI agents can discover each other, verify capabilities, and coordinate tasks without creating bottlenecks or security vulnerabilities. The project focuses on the technical infrastructure and the governance frameworks \footnote{\url{https://nanda.media.mit.edu/}}.

\subsection{Protocols}

In this section, we present protocols for agentic systems.
Given the focus on the Agentic Web, the scope of this work will cover general-purpose protocols.
\textbf{General-purpose protocols} focus on supporting multiple agents and contexts via a unified interface, while \textbf{domain-specific protocols} focus on optimizing for specific use cases and contexts \cite{Yang2025b}.
Moreover, it is important to contrast agent protocols from context protocols.
\textbf{Inter-agent protocols} are referred as being the ones allowing components to initiate a task autonomously, while \textbf{context-oriented protocols} are the ones that support connection of tools and models \cite{Yang2025b}.

\subsubsection{Model Context Protocol (MCP)}
MCP is an open transport-agnostic context protocol that enables developers to build secure, two-way connections between their data sources and LLM-powered tools. The architecture allows developers to expose their data through MCP servers or build AI applications (MCP clients) that connect to these servers using JSON-RPC for communication \citep{Anthropic2024}.
Considering this aspect related to connection with tools, MCP has pre-built MCP servers for Google Drive, GitHub, Slack, databases, among others.
Protocol main components: \textit{Client, Server, Tools, Resources, Prompts, and Sampling}.

MCP is also promoting the development of new technologies for the Web to improve how agents perform tasks on the Web. An example is WebMCP, an open source JavaScript library that helps on how websites expose available tools in a MCP-integrated way. 
WebMCP provides a standardized way to register tools, prompts and resources so that agents know exactly how to interact with a website's features, improving reliability and task performance on the Agentic Web \cite{Google2026, WebMCP2026}. 
WebMCP API methods: \textit{registerTool}, \textit{registerPrompt}, and \textit{registerResource}.

\subsubsection{Agent-to-Agent Protocol (A2A)}
A2A is an open client-server protocol that enables AI agents to communicate and collaborate across different platforms and frameworks, leveraging existing Web infrastructure. It's designed to maximize the benefits of agentic AI by enabling true multi-agent scenarios using JSON-RPC for communication. 
Key Features include interoperability (language- and framework-agnostic), enterprise-grade security, and flexibility \citep{Google2024}. 
Protocol main components: \textit{Agent Card, Skill, Task, Message, and Artifact}.

\subsubsection{Agent Communication Protocol (ACP)}
ACP is an open client-server protocol for agent interoperability aiming to support different frameworks and infrastructures. It considers agent-agent communication through a standardized RESTful API that supports different modalities, (a)synchronous communication, streaming interactions, stateful/less operations, agent discovery, and long tasks. 
Given that it is message-based, the protocol is agnostic to internal implementations \cite{ACP2025}.
Protocol main components: \textit{Agent Detail, Discovery Mechanisms, Task Request, Message Structure, and Artifacts}.

\subsubsection{Agent Network Protocol (ANP)}
ANP is an open-source peer-to-peer communication protocol for agents, defining how agents connect with each other, building an open, secure, and efficient collaboration network for agents \footnote{https://github.com/agent-network-protocol/AgentNetworkProtocol}.
It is a standardized cross-domain protocol based on processes in place.
Protocol main components: \textit{Agent Identity, Host, Client, Server, and Resource}.

\subsubsection{Agora Protocol}
Agora is an open protocol that provides decentralized communication primitives designed for large-scale LLM agent ecosystems. It aims at supporting communication among heterogeneous, decentralized agents \citep{Agora2024}.
It translates natural language into structured protocols, serving as an intermediary layer that maps user's task to the specific protocols required by agents. \cite{Yang2025b}

\subsubsection{Agent Payments Protocol (AP2)}
AP2 is an open protocol aiming at compliant agent-led payments across platforms. The protocol can be used as an extension of the A2A and MCP. It involves two main components for transactions \textit{Intent Mandate} and \textit{Cart Mandate}, allowing the comparison of the initial human-led task (intent) and the product/server purchased by the agent (cart) \footnote{\url{https://github.com/google-agentic-commerce/AP2}}.


Finally, instead of a specific stack of protocols, \citet{Ehtesham2025} recommend a different approach, i.e., an adoption roadmap for protocols, starting with MCP for tool access, then ACP for multimodal messaging, then A2A for collaborative task execution, and, finally, extending to ANP for decentralized agent marketplaces.
The rationale proposed by the authors is that \textit{``no single protocol suffices across all contexts''} \cite{Ehtesham2025}.

\subsection{Agentic Frameworks}

Building MAS requires frameworks to manage communication and orchestration among specialized roles. Given that Python is the most popular language\footnote{\url{https://spectrum.ieee.org/top-programming-languages-2025}}, our work will focus on frameworks in Python. Appendix \ref{app:frameworks} (Table \ref{tab:framewors}) presents MAS frameworks with active open-source projects. 
From the list of frameworks studied, the LLM-agnostic popular frameworks with higher level of compatibility with built-in tools, allowing exploring different workflow paradigms, are:
LangChain \cite{LangChain2025}, 
CrewAI \cite{Crewai2025}, 
LangGraph \cite{LangGraph2025}, and
Semantic Kernel \cite{SemanticKernel2025}

\subsection{Benchmarks for Web Tasks}

To systematically evaluate agents, several benchmarks and environments have been proposed. Appendix \ref{app:benchmarks} (Table \ref{tab:benchmarks}) presents a list of benchmarks used in assessments involving agentic environments. In the context of Agentic Web for agent-as-user, benchmarks that align with tasks and tools available at the Web were proposed in works such as Mind2Web \cite{Deng2023}, WebArena \cite{Zhou2023}, and WebShop \cite{Yao2023}, detailed next.

\textbf{WebArena} provides a benchmark that aims at evaluating tasks that humans routinely perform on the Internet, encompassing the following domains: online shopping, discussion forums, collaborative development, and business content management.
The content populating these self-hosted websites was extracted from their real-world counterparts.
It covers 812 long-horizon web-based tasks, described as a high-level natural language intent \cite{Zhou2023}.

\textbf{WebShop} provides a benchmark comprising simulated e-commerce website environment with 1.18 million real-world products scraped from Amazon.com and over 12,000 crowd-sourced text instructions (prompts) for agents to find a given target product. It aims at providing challenges for natural language processing for agents, including compositional instructions query formulation, and acting on noisy text \cite{Yao2023}.

\textbf{Mind2Web} provides a benchmark that aims at supporting the development of generalist agents for the Web that follow language instructions to complete complex tasks on any website. It was built to encompass three main aspects (1) diverse domains, websites and tasks, (2) use real-world websites, and (3) broad spectrum of user interaction patterns.
Mind2Web encompasses over 2,000 tasks, from 137 websites, spanning 31 different application domains such as travel, information, service, shopping, and entertainment \cite{Deng2023}.

\subsection{MAS Evaluations}

Beyond benchmarks, context protocols and client-server communication has also been assessed in the literature. 
MCP Interviewer \cite{MCPInterviewer2025} was proposed as a tool to evaluate how MCP servers behave in multiple scenarios. Results suggest that MCP servers \cite{MCPInterviewer2025, Fourney2025}: 
(1) do not know which clients or models they are working with,
(2) differ in terms of handling larger contexts, 
(3) respond differently to common prompting patterns, and 
(4) face name collision in terms of tools available (755 collisions under the term ``search'').

In MAS, prompts tailored to agents, instilling good heuristics rather than rigid rules, results in better outcomes \cite{Anthropic2025}. Anthropic provides the following guidelines for prompting for MAS \cite{Anthropic2025}: 
(1) think like your agents, 
(2) teach the orchestrator how to delegate, 
(3) scale effort to query complexity, 
(4) tool design and selection are critical, 
(5) let agents improve themselves, 
(6) start wide, then narrow down, 
(7) guide the thinking process, and 
(8) parallel tool calling transforms speed and performance.
The following extracts from a research MAS showcase these guidelines applied to prompting for agents \footnote{\url{https://github.com/anthropics/claude-cookbooks/tree/main/patterns/agents/prompts}}:
\begin{itemize}
    \item \textbf{Lead researcher agent (Orchestrator):}\textit{ ``[...] As the lead research agent, your primary role is to coordinate, guide, and synthesize - NOT to conduct primary research yourself. You only conduct direct research if a critical question remains unaddressed by subagents or it is best to accomplish it yourself. Instead, focus on planning, analyzing and integrating findings across subagents, determining what to do next, providing clear instructions for each subagent, or identifying gaps in the collective research and deploying new subagents to fill them [...]''}
    \item \textbf{Research agent (Subagent):} \textit{``[...] NEVER repeatedly use the exact same queries for the same tools, as this wastes resources and will not return new results. Follow this process well to complete the task. Make sure to follow the description and investigate the best sources'' [...] ``when you are no longer finding new relevant information and results are not getting better, STOP using tools and instead compose your final report [...]''}
    \item \textbf{Citation agent (Subagent):} \textit{``[...] ONLY add citations where the source documents directly support claims in the text [...]''}
\end{itemize}

When performing evaluations against MAS, LLM-as-a-judge has shown to be an effective and scalable solution when the eval test cases have a clear answer \cite{Anthropic2025}. 
When evaluating MAS, long conversations management is also an increasing challenge as some agents engage in conversations spanning hundreds of turns, cases in which compression and memory mechanisms are required such as summarization of completed work, storage of essential information in external memory before proceeding to new tasks, and spawning of subagents with clean contexts \cite{Anthropic2025}.

\newpage

\section{Problem Definition}
\label{sec:problem_definition}

The formalization presented next aims at connecting the technology stack for agents with research questions posed in section \ref{sec:introduction} in a verifiable way. The metrics presented next allow for ranking technology stacks for agents and, hence, to identify promising directions in terms of technology for the Agentic Web.

First, let $\mathcal{A}=\{a_1,\dots,a_n\}$ denote a set of agents. Each agent instance $a$ runs in an environment that combines a set of technologies defined in this study as a technology stack (TS)
\begin{align*}
\text{TS} = (\mathrm{LLM}, \mathrm{AP}, \mathrm{IF} ), 
\end{align*}
where $\mathrm{LLM}$ is the model agents are using, $\mathrm{AP}$ is the agent protocol, and $\mathrm{IF}$ the implementation framework.

A prompt $P$ is a tuple $P=(T,C)$ where $T$ is a high-level instruction (task) and $C$ is a given initial context, which may be empty for initial prompts or even poorly defined prompts. 

An orchestrator (or planner agent) $a_{orch}$ may decompose $T$ into subtasks $\{t_1,\dots,t_k\}$ and dispatch prompts $p_i=(t_i,c_i)$ to subagents $a_{i}$. The orchestrator collects subagent outputs $o_{a_i}$ and returns a final answer $o_{orch}$.

The action space for an agent includes an LLM-powered language space $\mathcal{L}$ and output actions drawn from (a subset of) $\mathcal{L}$. Each agent $a$ produces a reasoning trace $a_t$; hence
\begin{align*}
a_{i,t1} &= TS(P(t_1, C)) \\ 
a_{i,t2} &= TS(P(t_1, a_{i,t1})) \\
a_{i,t3} &= TS(P(t_1, a_{i,t2})) \\
\dots\\
a_{i,tm} &= TS(P(t_1, a_{i,t(m-1)})) \\
o_{ai} &= TS(P(t_1, a_{i,tm}))
\end{align*}
where $a_t$ is the reasoning trace and $o_a$ is the output (which may be used as a context).

\subsection{Task Completion}

Let $\Phi(T, C, O)$ be a task-specific predicate indicating whether output $O$ satisfies the instructions in task $T$. As recommended in \cite{Anthropic2025,Kapoor2024}, we consider $\Phi$ to be realized either by: 
\begin{enumerate}
\item An LLM-as-a-judge with a rubric informing success/failure criteria, or 
\item Human assessment with an success/failure criteria.
\end{enumerate}
Given that the output for a given task is given by
\begin{align*}
\mathrm{o_{orch} = TS(P(T, \left\{ o_{a1}, ... , o_{an} \right\}))},
\end{align*}
we say a task instance $(T,C)$ is \emph{completed} under MAS technology stack $TS$ if
\begin{align*}
\mathrm{\Lambda}_{TS}(T,C) = \mathbb{I}\left\{\Phi(T,\ C, o_{orch}\big) = 1\right\}, 
\end{align*}
where $\mathbb{I}\{\cdot\}$ is the indicator function.

Consider a test set $\mathcal{D}=\{(T_j,C_j)\}_{j=1}^{|\mathcal{D}|}$, we define the Task Completion Rate (TCR) as
\begin{align}
\mathrm{TCR_{TS}}(\mathcal{D}) = \frac{1}{|\mathcal{D}|} \sum_{P \in \mathcal{D}}^{} \mathrm{\Lambda}_{TS}(T,C) \label{eq:tcr}
\end{align}



\newpage

\subsection{Efficiency}

Efficiency can be seen as the metric that combines high effectiveness with low use of resources. Hence, in the context of this work, we propose assessing efficiency as a multi-dimensional efficiency profile composed of: task completion, convergence time (or the time a MAS takes to answer a task), and verbosity (or tokens spent by a MAS to answer a task). 

Convergence time (CT) is then defined as the time taken by agents to converge and provide an answer for a given prompt. For a given task instance with prompt $P=(T,C)$, let $\mathrm{T}(\cdot)$ denote the time elapsed from the moment the orchestrator receives $P$ until the system it produces a final output $o_{orch}$.
\begin{align}
\mathrm{CT_{TS}}(\mathcal{D}) = \frac{1}{|\mathcal{D}|} \sum_{P \in \mathcal{D}}^{} {\mathrm{T}} \left\{ \mathrm{\Lambda}_{TS}(T,C) \right\}. \label{eq:convergence_time}
\end{align}




We now focus the efficiency formulation specifically for each TS. The goal is to quantify how choices in TS affect reasoning traces and the tokens generated.
In the context of GenAI, cost in terms of input and output tokens is an important metric for comparing solutions \cite{Kapoor2024}.
Hence, technology stack efficiency (E) takes into account internal reasoning and tokens used. Hence, verbosity (V) is defined in terms of average tokens spent per task:
\begin{align}
\mathrm{V}_{TS}(\mathcal{D}) \;=\; \frac{1}{\mathcal{|D|}} \sum_{P \in \mathcal{D}}^{}
\sum_{a \in \mathcal{A}}^{}
||a_{it}|| + ||o_{ai}||
\label{eq:verbosity}
\end{align}

Where $||a_{it}||$ represents the total number of tokens in the reasoning trace for agent $a_i$ and $||o_{ai}||$ the total number of tokens of the output generated by an agent $a_i$.
Agents may generate output $o_{ai}$ with no reasoning steps, i.e., $||a_{it}|| = 0$.

With these components in hand, efficiency can be seen as a multi-dimensional efficiency profile considering task completion, convergence time, and verbosity of agents, represented respectively by ${TCR}_{TS}$, ${CT}_{TS}$, and ${V}_{TS}$.

\begin{align}
\mathrm{E}_{TS}(\mathcal{D}) \;=\; ({TCR}_{TS}(\mathcal{D}), {CT}_{TS}(\mathcal{D}), {V}_{TS}(\mathcal{D}))
\label{eq:efficiency}
\end{align}


\subsection{Tool Use}
Given that agents have freedom to select tools given a set of provided tools, we need to assess how they perform under this condition. Hence, let $a_{i,t_m}$ denote the reasoning trace of agent $a_i$ at task $t_m$, where:
\[
a_{i,t_m} = \mathrm{TS}(P(t, a_{i,t_{m-1}})).
\]

A reasoning trace $a_{i,t_m}$ may trigger a tool call. Let $\mathcal{U}_{TS}$ be the set of all tool calls induced by agents  performing tasks in $\mathcal{D}$. Each tool call $u \in \mathcal{U}_{TS}$ is associated with:
\begin{itemize}
    \item A tool identifier $\theta(u)$, and
    \item An execution outcome $\delta(u) \in \{0,1\}$.
\end{itemize}

Let $\mathcal{T}$ be the set of available tools. We define the tool use success indicator:
\[
tsi(u) =
\begin{cases}
1 & \text{if } \theta(u) \in \mathcal{T} \text{ and execution succeeds} \\
0 & \text{otherwise (e.g., unsuccessful, hallucination, etc.)}
\end{cases}
\]

Hence, we define tool use success rate (TUSR) as the success rate over all tool calls:
\begin{align}
\mathrm{TUSR_{TS}}(\mathcal{D}) =
\frac{1}{|\mathcal{U}_{TS}|} \sum_{u \in \mathcal{U}_{TS}} tsi(u).
\label{eq:tool_use}
\end{align}

\newpage

\subsection{Social Values}

Let $SV=\{sv_1, sv_2, \dots, sv_x\}$ denote a set of social values, e.g., trust, compliance, integrity, and safety. Each prompt $P$ consists of sentences $s \in P$, and a classifier assigns a $sv_i \in SV$ to each $s$. For an agent $a_i$, define its set of used social values (SV) for a given prompt as
\begin{align}
\mathrm{SV_{i}(P)} = \bigcup_{s \in P} \kappa(s),
\label{eq:values}
\end{align}

Where $P$ is the prompts handled by $a_i$ and $\kappa$ a classifier to retrieve a social value per sentence.
We now formulate the value stability from prompt, internal reasoning steps, up to final output. Considering agents consistency, internal value stability (IVS) can be measured as a Pairwise Jaccard-based formula: 



\begin{align}
& x_0 = P, \quad x_j = a_i{t_j} \;\; (1 \le j \le m), \quad x_{m+1} = o_{a_i}, \quad S_j = SV(x_j) \notag \\[6pt]
& \mathrm{IVS}(a_i) = \frac{2}{(m+1)(m+2)} \sum_{j=0}^{m} \sum_{k=j+1}^{m+1} \frac{\left| S_j \cap S_k \right|}{\left| S_j \cup S_k \right|}
\label{eq:internal_value_stability}
\end{align}

Where $a_{t1} \dots a_{tm}$ represent internal reasoning steps for an agent. P represents the initial prompt and $o_{a_{i}}$ the final output provided by the orchestrator. Hence, internal value stability is used to compute how social values change through the process for a single task.
Moreover, the social value stability (SVS) score for a technology stack TS is defined as:

\begin{align}
\mathrm{SVS}_{TS}(\mathcal{D}) \;=\; \frac{1}{\mathcal{|D|}} \sum_{P \in \mathcal{D}}^{} IVS(a_i)
\label{eq:stack_value_stability}
\end{align}

Where $SVS$ is a score of social value stability for a technology stack through the test set $\mathcal{D}$.

\section{Experiment Setup}
\label{sec:experiment_setup}

In this section we detail the materials considered in this study, including models, protocols, benchmark, testing framework (MASTIF), and tools provided to agents. Then, we detail the procedure and result analysis, based on the problem definition proposed in Section \ref{sec:problem_definition}.

\subsection{Materials}

\subsubsection{Protocols}

The rationale on considering MCP accompanied by agent protocols resides in the fact that the latter carry structured JSON and artifacts while the former will be responsible for routing tool calls through. Hence, MCP focuses on standardizing and how AI models and agents interact, in a structured way, with tools, APIs, data sources and any other external resource, while agent protocols aim on standardizing how independent agents interact with each other to discover, negotiate, delegate, share tasks, exchange contexts, or deliver results \cite{A2AProtocol2025}.
Given this context and the announcement that ACP is being merged with A2A \cite{Blair2025}, the selected protocols for this study are MCP and A2A (Table \ref{tab:experiment_overview}).
Protocols were compared in terms of overhead applied to the communication among agents. The standard condition will be performed in terms of messages without protocols' directives in exchanged messages.

\subsubsection{Frameworks}

Considering the list of frameworks from the literature (Appendix \ref{app:frameworks}, Table \ref{tab:framewors}), the LLM-agnostic popular frameworks with  higher level of compatibility and easy integration with the protocols studied are:
LangGraph \cite{LangGraph2025},
LangChain \cite{LangChain2025}, 
CrewAI \cite{Crewai2025}, and 
Semantic Kernel \cite{SemanticKernel2025}
(Table \ref{tab:experiment_overview}).

\subsubsection{Benchmark}

In terms of benchmark, considering the ones present in the literature (Appendix~\ref{app:benchmarks}, Table \ref{tab:benchmarks}), the one that provides support for assessing generalist agents for the Web for real-world websites, considering multiple domains, is the Mind2Web \cite{Deng2023}. The rationale was to consider a benchmark for agents to perform tasks involving e-commerce, travel planning, and enterprise knowledge assistants \cite{Yang2025}.
In this experiment, a stratified sample of 300 tasks was considered to encompass different websites and domains. 

Examples of high-level tasks present in the Mind2Web \cite{Deng2023} include: 

\begin{itemize}
    \item \textit{``Favorite a shared room in Belo Horizonte costing up to \$100 to stay in a weekend in May.''}
    \item \textit{``Find the results of the most recent NFL games.''}
    \item \textit{``Add a set of sonoma bath towels to the cart and apply a coupon code for free shipping.''}
    \item \textit{``Find hotel deals in Las Vegas for four adults starting on May 17 and ending on May 20, and if deal is not available, set an alert for the same.''}
    \item \textit{``Browse certified fresh sci-fi films streaming at home.''}
\end{itemize}

\subsubsection{Large language Models}

The following models hosted on Hugging Face were selected based on capabilities for tool use and agent reasoning \cite{Jiang2023, Deepseekai2025, MetaAI2025, QwenTeam2024}:

\begin{itemize}
    \item Mistral-7B-Instruct-v0.1: Baseline for instruction following and efficiency.
    \item Qwen2.5-7B: Evaluating tool-use accuracy and structured output.
    \item Llama-4-Scout-17B-16E-Instruct: Long-context memory and multimodal reasoning.
\end{itemize}

Judge model selected is gpt-4o-mini in a way to consider a different family model to judge MAS responses. The rationale for this decision aims at preventing \textit{preference leakage} as models from the same family tend to prefer outputs similar to their own distributions given shared LLM baselines and benchmarks \cite{Li2025}, while controlling for costs \cite{Kapoor2024} by selecting a cost-benefit model \cite{OpenAI2024}. Table \ref{tab:experiment_overview} summarizes the technology stack considered in this paper. Appendix \ref{app:judge_rubric} details judge model rubric used to assess MAS answers.

The rationale for selecting these models and judge was to balance cost and tool use in a MAS \cite{Kapoor2024}.







\subsubsection{MAS Test Framework}

\begin{figure*}[h]
  \centering
  \includegraphics[width=\linewidth]{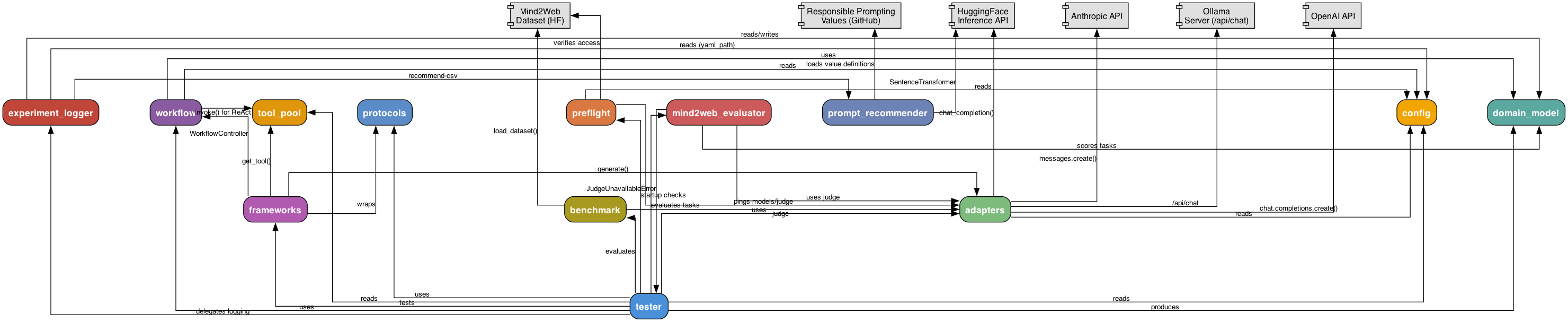}
  \caption{Overview of Multi-Agent System TestIng Framework (MASTIF).}
  \Description{Component diagram of the MASTIF (Multi-Agent System TestIng Framework), which combines frameworks, protocols, and frameworks.}
  \label{fig:mastif_component_diagram}
\end{figure*}

In order to test different GenAI models and agentic frameworks, a testing framework was developed, MASTIF (Multi-Agent System TestIng Framework). It is open source and available on GitHub \footnote{\url{https://github.com/cewebbr/mastif}}.
The goal of MASTIF is to support the specification of multiple tasks, a prompt template, connection with different inference providers (i.e., HuggingFace, OpenAI, Anthropic, and Ollama), and testing of multiple models and implementation frameworks. The testing framework logs latency, token usage, intermediate reasoning steps, and other metrics to enable the assessment of MAS on the Web. Figure \ref{fig:mastif_component_diagram} depicts MASTIF's main components.

MASTIF offers the following set of tools to be used by agents. The tools include 13 built-in tools compatible with the frameworks selected and two custom tools (i.e., $web\_interaction$ and $keyboard\_interaction$) that support interactivity with the Web, as needed for Mind2web tasks. The provided tools involve core \& utility tools, knowledge tools, computation \& parsing tools, content extraction tools, and interaction tools, as follows:

\begin{itemize}
\item $web\_search$: DuckDuckGo web search (via duckduckgo search).
\item $web\_browser$: Playwright headless browser navigation.
\item $wikipedia$: Wikipedia lookup (via wikipedia).
\item $arxiv$: Academic paper search (via arxiv).
\item $python\_repl$: Sandboxed Python code execution (via Restricted Python).
\item $requests\_get$: Simple HTTP GET requests (via requests)
\item $beautifulsoup\_scraper$: HTML structure extraction (via beautifulsoup4).
\item $pdf\_reader$: PDF text extraction (via pypdf).
\item $datetime$: Current date and time (stdlib).
\item $json\_parser$: JSON string parsing and querying (stdlib).
\item $youtube\_transcript$: YouTube video transcript retrieval (via youtube-transcript-api).
\item $sympy$: Symbolic math and logic evaluation (via sympy).
\item $web\_interaction$: Interactive browser tool for multi-step web interactions (via Playwright).
\item $keyboard\_interaction$: Keyboard-driven browser interaction optimized for accessibility-style tasks (via Playwright).
\end{itemize}

Following good practices for describing tools to agents \cite{Aizawa2025}, the following tool descriptions were provided to agents to promote tool usage: 

\begin{itemize}
\item $web\_search$: Use this tool when you need to find information that is not already known or may be outdated. This includes current events, recent facts, or unknown topics. Input should be a clear search query. Do NOT use this for simple reasoning or when the answer is already known.  
\item $web\_browser$: Use this tool to open and read the content of a specific webpage when you already have a URL. Useful after performing a search. Do NOT use this for searching—use $web\_search$ first.  
\item $requests\_get$: Use this to retrieve raw content from a specific URL (API, webpage, or endpoint). Prefer this for structured or programmatic access. Do NOT use for general browsing or search.  
\item $beautifulsoup\_scraper$: Use this to extract structured information from HTML content of a webpage. Typically used after fetching a page via $web\_browser$ or $requests\_get$. Do NOT use if you only need raw text.  
\item $datetime$: Use this when you need the current date or time. Do NOT guess or fabricate current timestamps.  
\item $wikipedia$: Use this for general knowledge lookups on well-known topics, people, or concepts. Faster and more reliable than $web
_search$ for encyclopedic content. Do NOT use for recent or niche information.  
\item $arxiv$: Use this to find academic papers in technical or scientific domains. Input should be a research topic or keywords. Do NOT use for general knowledge or non-academic queries.  
\item $pubmed$: Use this for biomedical or clinical research queries. Input should be medical terms or research questions. Do NOT use for general health advice or non-scientific queries.  
\item $python\_repl$: Use this for calculations, data processing, or executing code. Required for precise arithmetic, transformations, or simulations. Do NOT use for simple math that can be done mentally.  
\item $sympy$: Use this for symbolic mathematics such as algebra, equation solving, or simplification. Prefer this over $python\_repl$ for symbolic operations.  
\item $json\_parser$: Use this to parse, extract, or query structured JSON data. Input should be a JSON string and query instructions. Do NOT use for plain text.  
\item $pdf\_reader$: Use this to extract text from PDF files or PDF URLs. Required when the content is not directly readable as text. Do NOT use for HTML pages.  
\item $youtube\_transcript$: Use this to retrieve the transcript of a YouTube video. Input should be a video URL. Do NOT use if you only need metadata or summaries.  
\item $web\_interaction$: Use this to interact with webpages in multi-step tasks. Includes clicking elements, filling forms, navigating pages, and extracting results. Required for tasks involving user workflows on websites. Do NOT use for simple retrieval—use $web\_search$ or $web\_browser$ instead.  
\item $keyboard\_interaction$: Use this for keyboard-based navigation of web interfaces. Includes tabbing between elements, pressing enter, typing text, and triggering shortcuts. Prefer this when interaction must simulate accessibility or non-mouse workflows. Do NOT use for static content extraction.
\end{itemize}

\subsubsection{Social Value Classifier}

The social value classifier considered in this assessment will be based on the work from \citet{Santana2025}, reusing the social value classification based on similarity with a corpus of human-curated prompts and social values, powered by light weight the sentence transformers such as \textit{all-minilm-l6-v2}. The system retrieves a set of social value based on a given text input. The classifier thus retrieves a class (social value) for each sentence in the content provided, be it a task, prompt, agent reasoning step, or final answer.
The recommendation system is open source and can be tested via its cookbook notebooks, available on Github\footnote{\url{https://github.com/IBM/responsible-prompting-api/tree/main/cookbook}}.
This approach will be used to assess Equation \ref{eq:stack_value_stability}.

\subsection{Experiment Design and Procedure}

Based on the selected benchmark and testing framework, the following set of models, frameworks, protocols, and tasks will be considered (Table \ref{tab:experiment_overview}).
\begin{align}
\mathrm{Tests = |models| \times |frameworks| \times |protocols| \times |tasks|}
\label{eq:tests}
\end{align}

Hence, based on Equation \ref{eq:tests} and stratified sampling \cite{Ghojogh2020} of 300 out of 2,350 Mind2Web tasks, a total of 
$3 \times 4 \times 3 \times 300 = 10,800$ 
task executions will be evaluated.
Instead of simple random sampling, stratification was employed to allow selecting tasks from different domains and tasks of different complexities \cite{Ghojogh2020}.
Additional parameters considered in the experiment include: 

\begin{itemize}
    \item $temperature$: 0.7, aiming at diverse search interpretation; lower temperature tends to fix behavior on how agents use tools and higher temperature tends to allow different behaviors in terms of how to perform a task, tool use, including tool call hallucinations.
    \item $max\_tokens$: 1024, to support cumulative \\ \{research\_results\}.
    \item $max\_steps$: 3, to allow multiple agent loops if needed.
    \item $max\_tool_rounds$: 3, to allow agents to use one or more tools multiple times.
    \item $timeout$: 100 seconds, increased timeout to handle automation restrictions placed by some search engines.
    \item $sampling\_seed$: 20260504.
\end{itemize}

Beyond these parameters, MASTIF also supports defining the agentic workflow (linear and hierarchical), including nodes with and without loops, each node with its configurable prompt.
Hence, considering the format of tasks in Mind2Web \cite{Deng2023}, the workflow used in this experiment considers 4 nodes in the following way: 

\begin{enumerate}
    \item Navigating;
    \item Observing the webpage at hand (loop node);
    \item Interaction with the webpage at hand (loop node), and;
    \item Final reporting.
\end{enumerate}

Open models and tasks used HuggingFace API while judge model used OpenAI API. 
All the specifications for this experiments are detailed in the 
YAML\footnote{https://github.com/cewebbr/mastif/tree/main/experiments/tweb2026.yaml} file ingested by the testing framework,  prompts\footnote{https://github.com/cewebbr/mastif/tree/main/prompts/mind2web/} considered in the workflow, and resulting logs\footnote{https://github.com/cewebbr/mastif/tree/main/logs/} are all available on GitHub.




\begin{table}
  \caption{Experiment setup overview.}
  \label{tab:experiment_overview}
  \begin{tabular}{p{2cm}p{5.5cm}}
    \toprule
    Category & $TS_{model, protocol, framework}$ \\ 
    \midrule
    Models & 
    $\bullet$ Mistral-7B-Instruct-v0.1 \cite{Jiang2023} \newline
    $\bullet$ Qwen2.5-7B \cite{QwenTeam2024} \newline
    $\bullet$ Llama-4-Scout-17B-16E-Instruct \cite{MetaAI2025}
    \\ \midrule
    Judge model & 
    $\bullet$ GPT-4o-mini \cite{OpenAI2024}
    \\ \midrule
    Protocols & 
    $\bullet$ Standard call \newline
    $\bullet$ A2A \cite{A2AProtocol2025} \newline
    $\bullet$ MCP \cite{Anthropic2024}
    \\ \midrule
    Frameworks & 
    $\bullet$ LangGraph \cite{LangGraph2025} \newline
    $\bullet$ LangChain \cite{LangChain2025} \newline
    $\bullet$ CrewAI \cite{Crewai2025} \newline
    $\bullet$ Semantic Kernel \cite{SemanticKernel2025}
    \\ \midrule
    Benchmark & 
    $\bullet$ Mind2Web \cite{Deng2023}
    \\
  \bottomrule
\end{tabular}
\end{table}

\subsection{Result Analysis}

Result analysis will aim at comparing different Agentic Web technology stacks ($TS$) in terms of the following metrics, detailed in Section \ref{sec:problem_definition}:

\begin{itemize}
\item $TCR_{TS}$: task completion rate (Equation \ref{eq:tcr}),
\item $CT_{TS}$: convergence time (Equation \ref{eq:convergence_time}),
\item $V_{TS}$: verbosity (Equation \ref{eq:verbosity}),
\item $E_{TS}$: efficiency (Equation \ref{eq:efficiency}) plus a Pareto frontier analysis,
\item $TUSR_{TS}$ tool use (Equation \ref{eq:tool_use}), and
\item $SVS_{TS}$: social values stability (Equation \ref{eq:stack_value_stability}).
\end{itemize}

Shapiro-Wilk will be used to test for normal distribution and then, depending on the result, ANOVA or Kruskal-Wallis plus Dunn's Post-hoc tests for assessing significant differences considering models, protocols, and frameworks

\newpage

\section{Results}
\label{sec:results}

In this section we present results for the metrics defined in Section \ref{sec:problem_definition}. Shapiro-Wilk statistic was computed for all metrics detailed next and results suggest that they do not follow a normal distribution (p < 0.001). Hence, Kruskal-Wallis test was employed to assess significant differences considering models, protocols, and frameworks. Moreover, Dunn's Post-hoc test was be used to identify pairwise significant differences ($\alpha=0.05$)

\subsection{Task Completion Rate}

Task completion rate was computed according to Equation \ref{eq:tcr} and processed by the judge model according to the rubric detailed in Appendix \ref{app:judge_rubric}. 

Kruskal-Wallis test for task completion rate suggests that there are significant differences for model (H-statistic = 1203.263, $p-value$ < 0.001), protocol (H-statistic = 428.770, $p-value$ < 0.001), and framework (H-statistic = 6039.33, $p-value$ < 0.001).

Dunn's Post-hoc suggests significant pairwise differences for model when assessing task completing rates as follows:
\begin{itemize}
\item Llama-4-Scout-17B-16E vs. Mistral-7B ($p-value$ < 0.001) and
\item Llama-4-Scout-17B-16E vs. Qwen2.5-7B ($p-value$ < 0.001).
\end{itemize}

Dunn's Post-hoc suggests significant pairwise differences for protocol when assessing task completing rates as follows:
\begin{itemize}
\item A2A vs. MCP ($p-value$ < 0.001), 
\item A2A vs. standard ($p-value$ < 0.001), and
\item MCP vs. standard ($p-value$ < 0.001).
\end{itemize}

Dunn's Post-hoc suggests significant pairwise differences for framework when assessing task completing rates as follows:
\begin{itemize}
\item CrewAI vs. SemanticKernel ($p-value$ < 0.001),
\item LangChain vs. SemanticKernel ($p-value$ < 0.001), and
\item LangGraph vs. SemanticKernel: $p-value$ < 0.001).
\end{itemize}

\begin{figure*}[h]
  \centering
  \includegraphics[width=\linewidth]{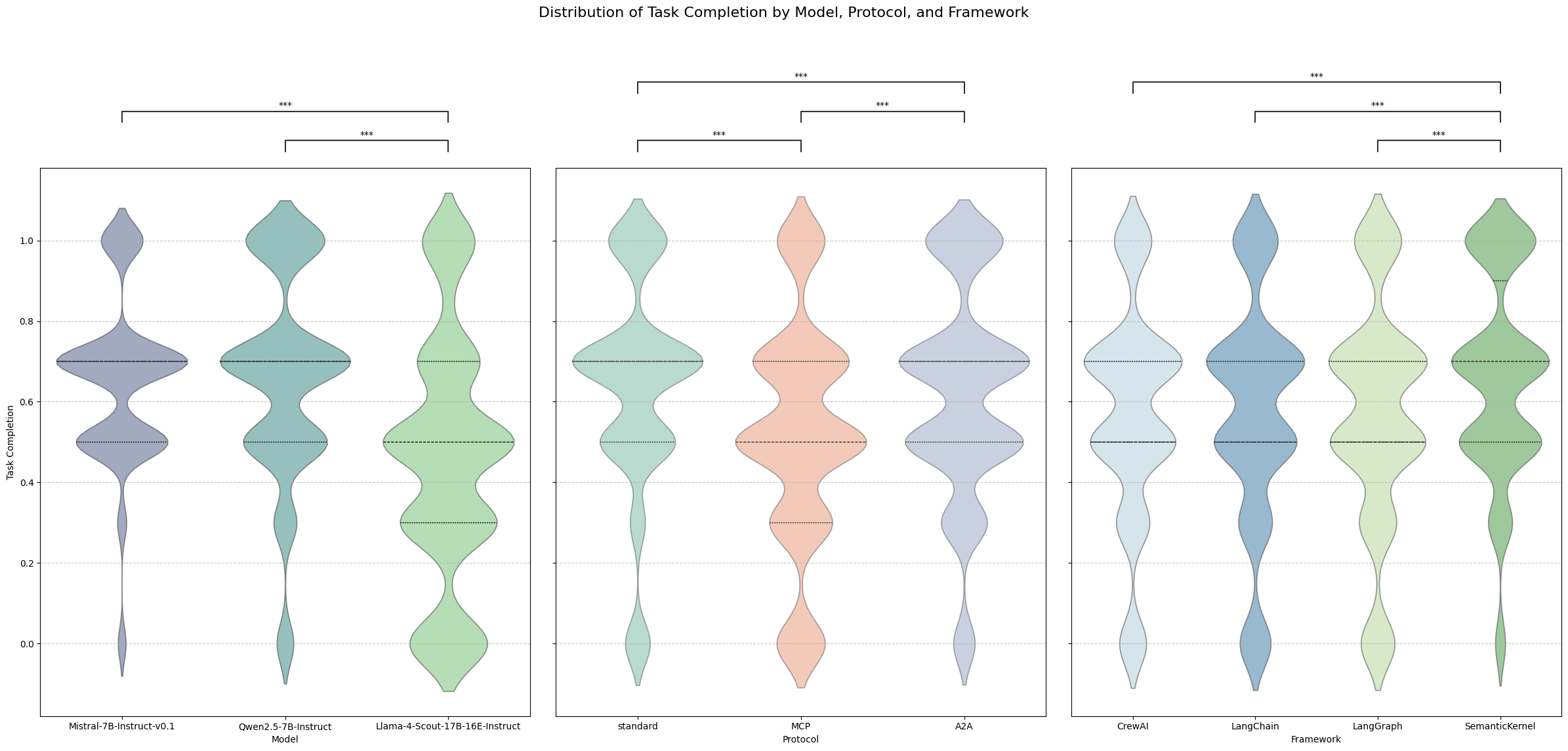}
  \caption{Task completion scores by model, protocol, and framework. Dunn's Post-hoc pairwise significant differences are shown according to the following coding: *** for $p-value$ < 0.001, ** for $p-value$ < 0.01, and * for $p-value$ < 0.05.}
  \Description{Task completion scores by model, protocol, and framework.}
  \label{fig:task_completion}
\end{figure*}

Figure \ref{fig:task_completion} shows a violin plot grouping task completion scores by model, protocol, and framework.

Table \ref{tab:tcr} shows mean scores obtained in technology stacks studied. Moreover, given that we also required judge model to describe its rationale and assessment of task understanding and how reasoning adheres to the task, we report additional scores retrieved by the model in terms of task understanding ($TU_{TS}$) and adherence ($TA_{TS}$), all according to the judge's provided rubric (Appendix~\ref{app:judge_rubric}).

Hence, results suggest, under the test conditions, that the best-performing technology stack aiming at task completion score should consider: \textbf{Qwen2.5-7B} (mean rank=6095.99), \textbf{standard} (mean rank=5898.66), and \textbf{SemanticKernel} (mean rank=6039.33).

\begin{table*}
\caption{Average task understanding ($TU_{TS}$), adherence ($TA_{TS}$), and task completion ($TCR_{TS}$) for different technology stacks, as provided by judge model given rubric detailed in Appendix \ref{app:judge_rubric}.}
\label{tab:tcr}
\begin{tabular}{lrrr}
\toprule

$TS_{model, protocol, framework}$ & $TU_{TS}$ & $TA_{TS}$ & $TCR_{TS}$ ($\sigma$) \\

\midrule

Mistral-7B, standard, CrewAI & \cellcolor[HTML]{9090FF} 0.910 & \cellcolor[HTML]{7777FF} 0.969 & \cellcolor[HTML]{FBFBFF} 0.667 (0.186) \\
Mistral-7B, MCP, CrewAI & \cellcolor[HTML]{9C9CFF} 0.883 & \cellcolor[HTML]{8080FF} 0.947 & \cellcolor[HTML]{FFFBFB} 0.633 (0.222) \\ 
Mistral-7B, A2A, CrewAI & \cellcolor[HTML]{9898FF} 0.893 & \cellcolor[HTML]{7E7EFF} 0.952 & \cellcolor[HTML]{FFFAFA} 0.628 (0.196) \\ \hline

Mistral-7B, standard, LangChain & \cellcolor[HTML]{8A8AFF} 0.925 & \cellcolor[HTML]{7070FF} 0.983 & \cellcolor[HTML]{F5F5FF} \textbf{0.680} (0.207) \\
Mistral-7B, MCP, LangChain & \cellcolor[HTML]{8D8DFF} 0.918 & \cellcolor[HTML]{6F6FFF} \textbf{0.987} & \cellcolor[HTML]{FEFEFF} 0.659 (0.199) \\ 
Mistral-7B, A2A, LangChain & \cellcolor[HTML]{9090FF} 0.910 & \cellcolor[HTML]{7777FF} 0.967 & \cellcolor[HTML]{FFFAFA} 0.631 (0.215) \\ \hline

Mistral-7B, standard, LangGraph & \cellcolor[HTML]{8787FF} 0.931 & \cellcolor[HTML]{7272FF} 0.980 & \cellcolor[HTML]{F8F8FF} 0.673 (0.187) \\
Mistral-7B, MCP, LangGraph & \cellcolor[HTML]{DBDBFF} 0.740 & \cellcolor[HTML]{CACAFF} 0.778 & \cellcolor[HTML]{FFF1F1} 0.579 (\textbf{\textit{0.181}}) \\ 
Mistral-7B, A2A, LangGraph & \cellcolor[HTML]{8A8AFF} 0.924 & \cellcolor[HTML]{7272FF} 0.979 & \cellcolor[HTML]{FAFAFF} 0.669 (0.223) \\ \hline

Mistral-7B, standard, Semantic Kernel & \cellcolor[HTML]{9B9BFF} 0.886 & \cellcolor[HTML]{7979FF} 0.963 & \cellcolor[HTML]{FFFBFB} 0.634 (0.222) \\
Mistral-7B, MCP, Semantic Kernel & \cellcolor[HTML]{9898FF} 0.892 & \cellcolor[HTML]{7F7FFF} 0.950 & \cellcolor[HTML]{FFFCFC} 0.642 (0.204) \\ 
Mistral-7B, A2A, Semantic Kernel & \cellcolor[HTML]{8686FF} \textbf{0.934} & \cellcolor[HTML]{7373FF} 0.977 & \cellcolor[HTML]{F9F9FF} 0.670 (0.211) \\ \hline

Qwen2.5-7B, standard, CrewAI & \cellcolor[HTML]{B3B3FF} 0.830 & \cellcolor[HTML]{ADADFF} 0.844 & \cellcolor[HTML]{F1F1FF} 0.689 (0.179) \\
Qwen2.5-7B, MCP, CrewAI & \cellcolor[HTML]{8282FF} 0.942 & \cellcolor[HTML]{7272FF} 0.980 & \cellcolor[HTML]{FFF1F1} 0.583 (0.269) \\
Qwen2.5-7B, A2A, CrewAI & \cellcolor[HTML]{8282FF} 0.942 & \cellcolor[HTML]{7070FF} 0.983 & \cellcolor[HTML]{F4F4FF} 0.682 (0.234) \\ \hline

Qwen2.5-7B, standard, LangChain & \cellcolor[HTML]{7474FF} 0.975 & \cellcolor[HTML]{6B6BFF} \textbf{0.995} & \cellcolor[HTML]{C3C3FF} \textbf{0.795} (\textbf{\textit{0.157}}) \\
Qwen2.5-7B, MCP, LangChain & \cellcolor[HTML]{7E7EFF} 0.952 & \cellcolor[HTML]{7474FF} 0.975 & \cellcolor[HTML]{FFE9E9} 0.537 (0.278) \\
Qwen2.5-7B, A2A, LangChain & \cellcolor[HTML]{8282FF} 0.942 & \cellcolor[HTML]{7070FF} 0.984 & \cellcolor[HTML]{F7F7FF} 0.676 (0.263) \\ \hline

Qwen2.5-7B, standard, LangGraph & \cellcolor[HTML]{7575FF} 0.973 & \cellcolor[HTML]{6E6EFF} 0.989 & \cellcolor[HTML]{C9C9FF} 0.780 (0.168) \\
Qwen2.5-7B, MCP, LangGraph & \cellcolor[HTML]{8484FF} 0.939 & \cellcolor[HTML]{7272FF} 0.979 & \cellcolor[HTML]{FFE4E4} 0.509 (0.268) \\
Qwen2.5-7B, A2A, LangGraph & \cellcolor[HTML]{8787FF} 0.932 & \cellcolor[HTML]{7373FF} 0.978 & \cellcolor[HTML]{F7F7FF} 0.676 (0.260) \\ \hline

Qwen2.5-7B, standard, Semantic Kernel & \cellcolor[HTML]{7373FF} \textbf{0.977} & \cellcolor[HTML]{6E6EFF} 0.988 & \cellcolor[HTML]{CFCFFF} 0.767 (0.170) \\
Qwen2.5-7B, MCP, Semantic Kernel & \cellcolor[HTML]{8282FF} 0.942 & \cellcolor[HTML]{7676FF} 0.970 & \cellcolor[HTML]{FFE4E4} 0.509 (0.279) \\
Qwen2.5-7B, A2A, Semantic Kernel & \cellcolor[HTML]{8080FF} 0.947 & \cellcolor[HTML]{7171FF} 0.982 & \cellcolor[HTML]{EFEFFF} 0.693 (0.267) \\ \hline

Llama-4-Scout-17B-16E, standard, CrewAI & \cellcolor[HTML]{6E6EFF} 0.988 & \cellcolor[HTML]{6D6DFF} 0.991 & \cellcolor[HTML]{FFD2D2} 0.412 (0.270) \\
Llama-4-Scout-17B-16E, MCP, CrewAI & \cellcolor[HTML]{FCFCFF} 0.664 & \cellcolor[HTML]{FFFEFE} 0.652 & \cellcolor[HTML]{FFC1C1} 0.322 (0.279) \\
Llama-4-Scout-17B-16E, A2A, CrewAI & \cellcolor[HTML]{D5D5FF} 0.754 & \cellcolor[HTML]{E3E3FF} 0.721 & \cellcolor[HTML]{FFE0E0} 0.489 (0.271) \\ \hline

Llama-4-Scout-17B-16E, standard, LangChain & \cellcolor[HTML]{7F7FFF} 0.949 & \cellcolor[HTML]{8080FF} 0.948 & \cellcolor[HTML]{FFCCCC} 0.379 (0.278) \\
Llama-4-Scout-17B-16E, MCP, LangChain & \cellcolor[HTML]{F9F9FF} 0.671 & \cellcolor[HTML]{FFF7F7} 0.611 & \cellcolor[HTML]{FFC3C3} 0.333 (0.264) \\
Llama-4-Scout-17B-16E, A2A, LangChain & \cellcolor[HTML]{D8D8FF} 0.747 & \cellcolor[HTML]{F2F2FF} 0.686 & \cellcolor[HTML]{FFDCDC} 0.464 (0.269) \\ \hline

Llama-4-Scout-17B-16E, standard, LangGraph & \cellcolor[HTML]{6D6DFF} \textbf{0.990} & \cellcolor[HTML]{6C6CFF} \textbf{0.992} & \cellcolor[HTML]{FFCBCB} 0.376 (0.283) \\
Llama-4-Scout-17B-16E, MCP, LangGraph & \cellcolor[HTML]{FFFEFE} 0.651 & \cellcolor[HTML]{FFF7F7} 0.615 & \cellcolor[HTML]{FFC0C0} 0.314 (0.282) \\
Llama-4-Scout-17B-16E, A2A, LangGraph & \cellcolor[HTML]{D5D5FF} 0.753 & \cellcolor[HTML]{E9E9FF} 0.707 & \cellcolor[HTML]{FFDFDF} 0.482 (0.274) \\ \hline

Llama-4-Scout-17B-16E, standard, Semantic Kernel & \cellcolor[HTML]{A8A8FF} 0.856 & \cellcolor[HTML]{AFAFFF} 0.839 & \cellcolor[HTML]{D2D2FF} \textbf{0.759} (\textbf{0.243}) \\
Llama-4-Scout-17B-16E, MCP, Semantic Kernel & \cellcolor[HTML]{BABAFF} 0.814 & \cellcolor[HTML]{E1E1FF} 0.725 & \cellcolor[HTML]{FFF5F5} 0.605 (0.278) \\
Llama-4-Scout-17B-16E, A2A, Semantic Kernel & \cellcolor[HTML]{ADADFF} 0.845 & \cellcolor[HTML]{CFCFFF} 0.767 & \cellcolor[HTML]{FFFFFF} 0.655 (0.266) \\

\bottomrule
\end{tabular}
\end{table*}

\clearpage

\subsection{Efficiency}

Efficiency was proposed as a multi-dimensional profile (Equation \ref{eq:efficiency}) that combines task completion score (Equation \ref{eq:tcr}) with convergence time (Equation \ref{eq:convergence_time}) and verbosity (Equation \ref{eq:verbosity}).

\subsubsection{Convergence Time}
Kruskal-Wallis test for \textbf{convergence time} suggests that there are significant differences for model (H-statistic = 81.538, $p-value$ < 0.001), protocol (H-statistic = 113.922, $p-value$ < 0.001), and framework (H-statistic = 807.034, $p-value$ < 0.001).

Dunn's Post-hoc suggests significant pairwise differences for model, when assessing convergence time, as follows:
\begin{itemize}
\item Llama-4-Scout-17B-16E vs. Mistral-7B ($p-value$ < 0.001),
\item Llama-4-Scout-17B-16E vs. Qwen2.5-7B ($p-value$ < 0.001) and
\item Mistral-7B vs. Qwen2.5-7B ($p-value$ < 0.001).
\end{itemize}

Dunn's Post-hoc suggests significant pairwise differences for protocol, when assessing convergence time, as follows:
\begin{itemize}
\item A2A vs. MCP ($p-value$ < 0.001) and
\item MCP vs. standard ($p-value$ < 0.001).
\end{itemize}

Dunn's Post-hoc suggests significant pairwise differences for framework, when assessing convergence time, as follows:
\begin{itemize}
\item CrewAI vs. LangChain ($p-value$ < 0.001),
\item CrewAI vs. LangGraph ($p-value$ < 0.001),
\item CrewAI vs. SemanticKernel ($p-value$ < 0.001),
\item LangChain vs. SemanticKernel ($p-value$ < 0.001), and
\item LangGraph vs. SemanticKernel ($p-value$ < 0.001).
\end{itemize}

\newpage

\begin{figure*}[t]
  \centering
  \includegraphics[width=\linewidth]{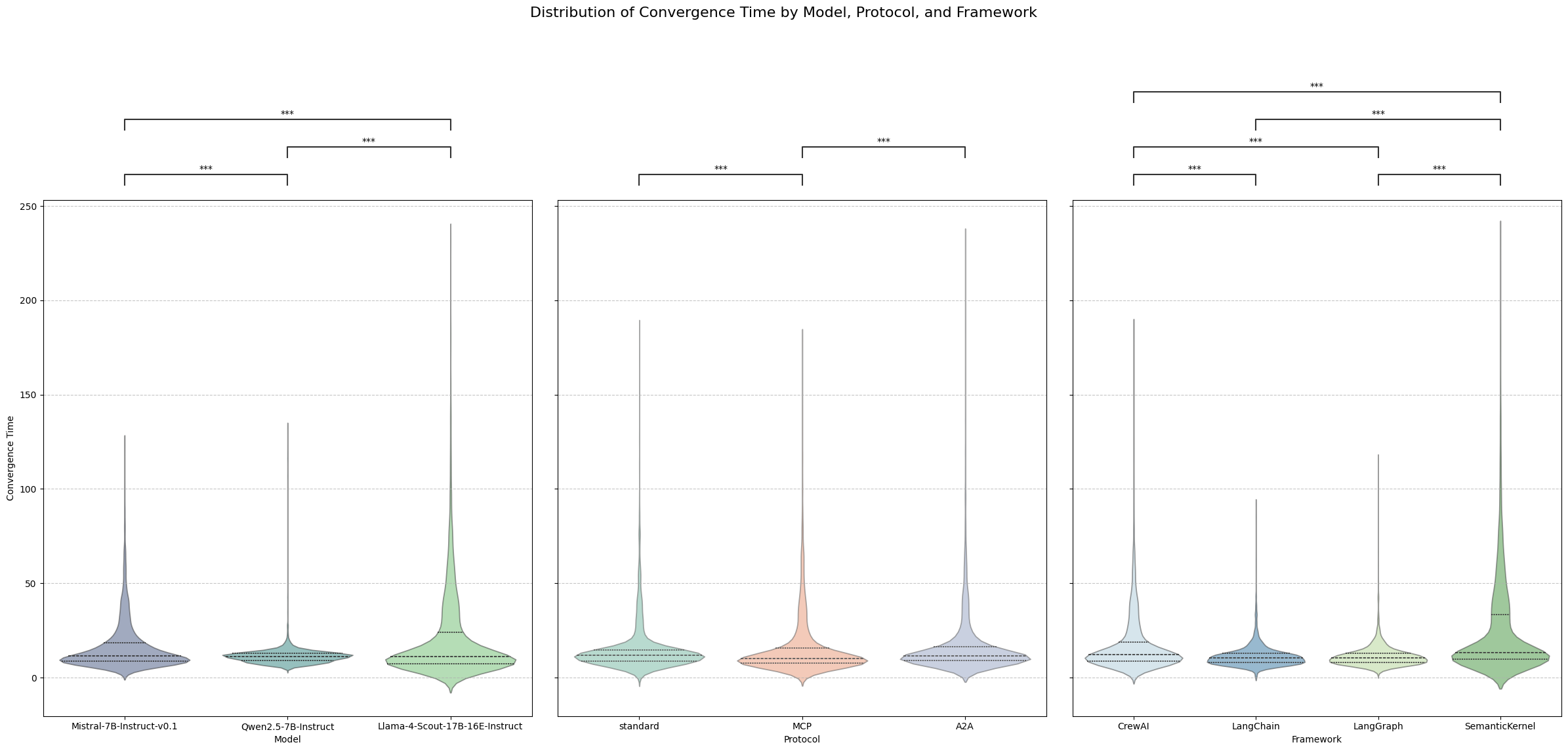}
  \caption{Convergence time in seconds grouped by model, protocol, and framework. Dunn's Post-hoc pairwise significant differences are shown according to the following coding: *** for $p-value$ < 0.001, ** for $p-value$ < 0.01, and * for $p-value$ < 0.05.}
  \Description{Convergence time by model, protocol, and framework.}
  \label{fig:convergence_time}
\end{figure*}

Figure \ref{fig:convergence_time} shows a violin plot grouping convergence time by model, protocol, and framework.
Considering significant differences found and Kruskal-Wallis mean ranks per group, results suggest that the best-performing technology stack aiming at lowest convergence time should combine:
\textbf{Qwen2.5-7B} (mean ranks=4974.57),
\textbf{MCP} (mean rank=4874.13), and
\textbf{LangChain} (mean rank=4510.52).

\subsubsection{Verbosity}
Kruskal-Wallis test for \textbf{verbosity} suggests that there are significant differences for model (H-statistic = 2483.926, $p-value$ < 0.001), protocol (H-statistic = 1156.463, $p-value$ < 0.001), and framework (H-statistic = 60.922, $p-value$ < 0.001).

Dunn's Post-hoc suggests significant pairwise differences for model, when assessing verbosity, as follows:
\begin{itemize}
\item Llama-4-Scout-17B-16E vs. Mistral-7B ($p-value$ < 0.001),
\item Llama-4-Scout-17B-16E vs. Qwen2.5-7B ($p-value$ < 0.001), and
\item Mistral-7B vs. Qwen2.5-7B ($p-value$ < 0.001).
\end{itemize}

Dunn's Post-hoc suggests significant pairwise differences for protocol, when assessing verbosity, as follows:
\begin{itemize}
\item A2A vs. MCP ($p-value$ < 0.001),
\item A2A vs. standard ($p-value$ < 0.001), and
\item MCP vs. standard ($p-value$ < 0.001).
\end{itemize}

Dunn's Post-hoc suggests significant pairwise differences for framework, when assessing verbosity, as follows:
\begin{itemize}
\item CrewAI vs. LangChain ($p-value$ < 0.001),
\item CrewAI vs. LangGraph ($p-value$ < 0.001), and
\item CrewAI vs. SemanticKernel ($p-value$ < 0.001).
\end{itemize}

\newpage

\begin{figure*}[h]
  \centering
  \includegraphics[width=\linewidth]{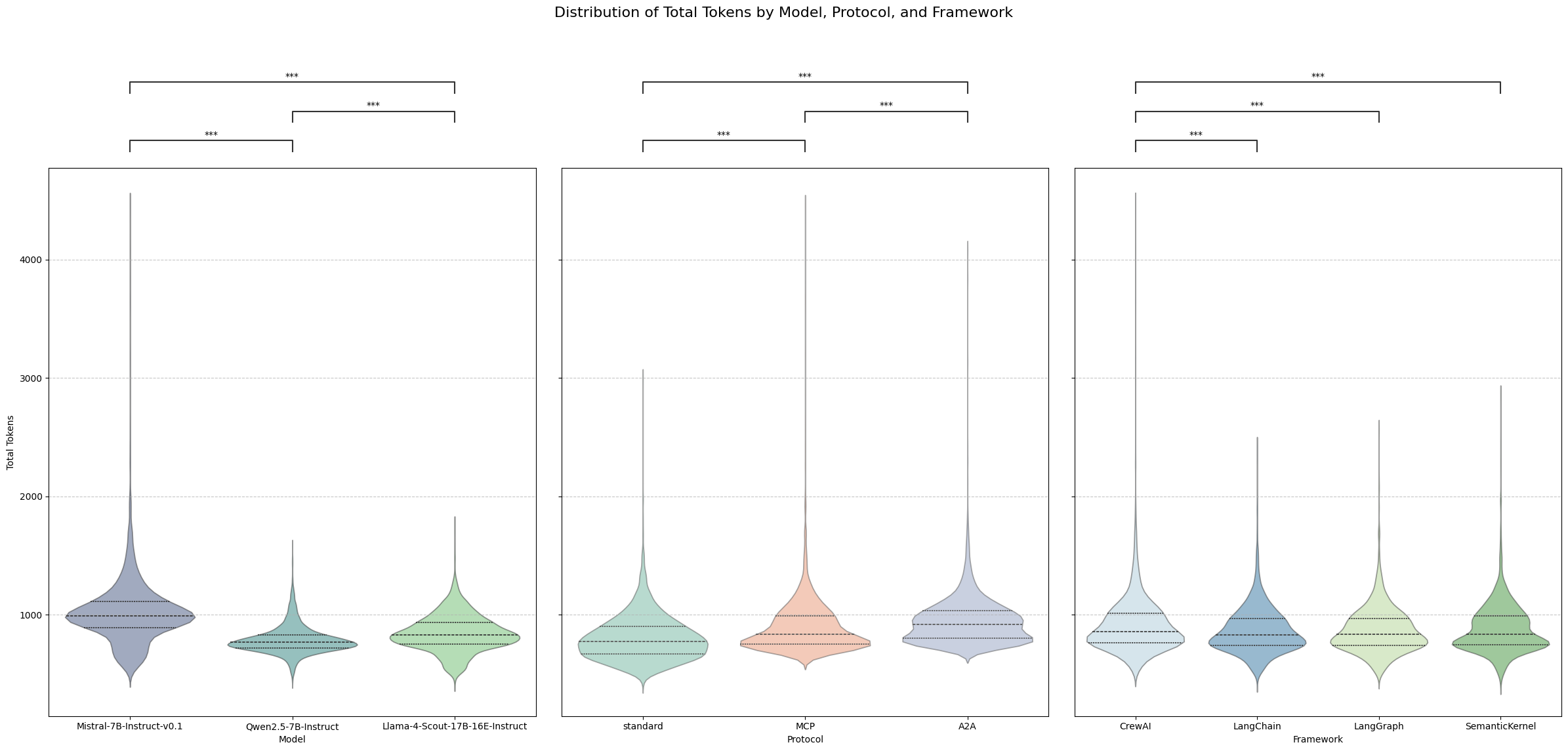}
  \caption{Verbosity in tokens grouped by model, protocol, and framework. Dunn's Post-hoc pairwise significant differences are shown according to the following coding: *** for $p-value$ < 0.001, ** for $p-value$ < 0.01, and * for $p-value$ < 0.05.}
  \Description{Verbosity time by model, protocol, and framework.}
  \label{fig:verbosity}
\end{figure*}

Figure \ref{fig:verbosity} shows a violin plot grouping verbosity (in tokens) by model, protocol, and framework.
Considering significant differences found and Kruskal-Wallis mean ranks per group, results suggest that the best-performing technology stack aiming at lowest verbosity should combine:
\textbf{Qwen2.5-7B} (mean ranks=3713.37),
\textbf{standard} (mean rank=4011.82), and
\textbf{LangChain} (mean rank=5113.11).

\subsubsection{Efficiency}

Bearing in mind that efficiency is being assessed as a multi-dimensional efficiency profile composed of task completion, convergence time, and verbosity of agents, results suggest that the technology stack composed of 
\textbf{Qwen2.5-7B}, 
\textbf{standard}, 
\textbf{LangChain} is a promising combination when aiming at high task completion rates while valuing low convergence time and low verbosity (Table \ref{tab:best_stack_summary}).

\begin{figure*}[h]
  \centering
  \includegraphics[width=\linewidth,trim={0 0 0 2cm},clip]{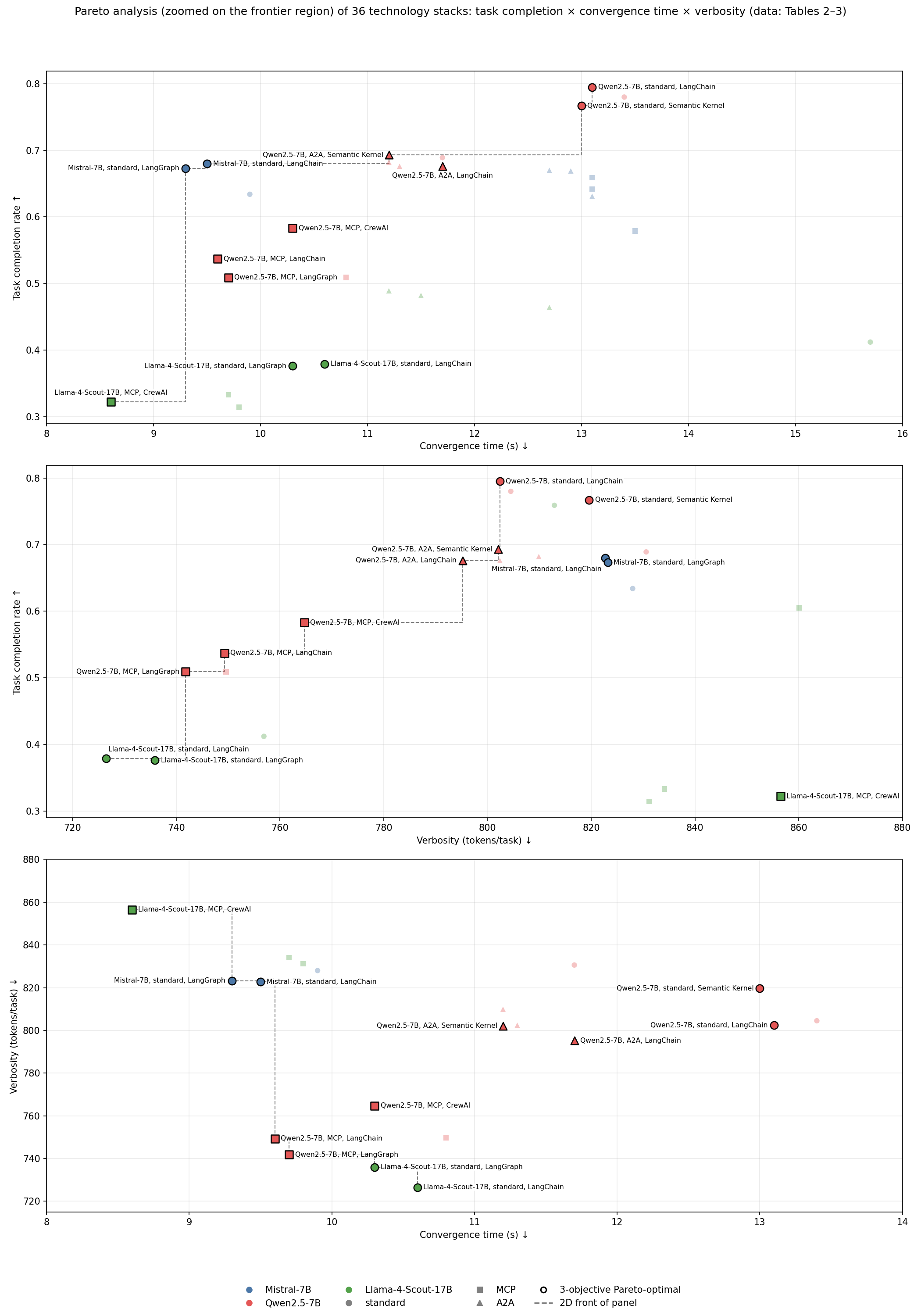}
  \caption{Pareto analysis zoomed on the frontier region of technology stacks combining task completion, verbosity, and convergence time.}
  \Description{Pareto analysis zoomed on the frontier region of technology stacks combining task completion, verbosity, and convergence time.}
  \label{fig:efficiency_pareto_frontier}
\end{figure*}

Figure \ref{fig:efficiency_pareto_frontier} shows a Pareto frontier analysis combining task completion, convergence time, and verbosity.
It is possible to identify different stack technologies that lead to slightly different results. 
Highest task completion scores obtained with \textbf{Qwen2.5-7B, standard, LangChain} and \textbf{Qwen2.5-7B, standard, Semantic Kernel}.
Technologies stacks that balance low convergence time with low verbosity are \textbf{Qwen2.5-7B, A2A, Semantic Kernel} and \textbf{Qwen2.5-7B, A2A, LangChain}.
Finally, Pareto frontier also shows that technology stacks with promising low convergence time and decent task completion are 
\textbf{Mistral-7B, standard, LangChain} and 
\textbf{Mistral-7B, standard, LangGraph}.

\begin{table*}
\caption{Mean convergence time ($\mathrm{CT_{TS}}$) per task in seconds and mean verbosity $\mathrm{V}_{TS}$ per task in tokens for different technology stacks.}
\label{tab:efficiency}
\begin{tabular}{lrr}
\toprule

$TS_{model, protocol, framework}$ & $\mathrm{CT_{TS}} (\sigma)$ & $\mathrm{V}_{TS} (\sigma)$ \\

\midrule

Mistral-7B, standard, CrewAI & \cellcolor[HTML]{B2B2FF} 20.8 (13.3) & \cellcolor[HTML]{E2E2FF} 912.4 (273.6) \\
Mistral-7B, MCP, CrewAI & \cellcolor[HTML]{FAFAFF} 32.8 (19.1) & \cellcolor[HTML]{FFC0C0} 1186.6 (405.5) \\
Mistral-7B, A2A, CrewAI & \cellcolor[HTML]{FAFAFF} 32.8 (17.2) & \cellcolor[HTML]{FFC3C3} 1176.6 (332.1) \\ \hline

Mistral-7B, standard, LangChain & \cellcolor[HTML]{6E6EFF} 9.5 (3.2) & \cellcolor[HTML]{A8A8FF} \textbf{822.7} (212.5) \\
Mistral-7B, MCP, LangChain & \cellcolor[HTML]{8484FF} 13.1 (6.9) & \cellcolor[HTML]{FFDBDB} 1088.0 (210.1) \\
Mistral-7B, A2A, LangChain & \cellcolor[HTML]{8484FF} 13.1 (7.8) & \cellcolor[HTML]{FFE3E3} 1059.9 (177.8) \\ \hline

Mistral-7B, standard, LangGraph & \cellcolor[HTML]{6D6DFF} \textbf{9.3} (\textbf{\textit{2.8}}) & \cellcolor[HTML]{A8A8FF} 823.2 (210.5) \\
Mistral-7B, MCP, LangGraph & \cellcolor[HTML]{8686FF} 13.5 (7.6) & \cellcolor[HTML]{FFE2E2} 1061.5 (200.8) \\
Mistral-7B, A2A, LangGraph & \cellcolor[HTML]{8383FF} 12.9 (7.3) & \cellcolor[HTML]{FFDDDD} 1079.6 (194.7) \\ \hline

Mistral-7B, standard, Semantic Kernel & \cellcolor[HTML]{7171FF} 9.9 (3.7) & \cellcolor[HTML]{ABABFF} 828.0 (212.0) \\
Mistral-7B, MCP, Semantic Kernel & \cellcolor[HTML]{8484FF} 13.1 (8.4) & \cellcolor[HTML]{FFDDDD} 1082.7 (241.7) \\
Mistral-7B, A2A, Semantic Kernel & \cellcolor[HTML]{8282FF} 12.7 (6.3) & \cellcolor[HTML]{FFE1E1} 1065.2 (\textbf{\textit{174.0}}) \\ \hline

Qwen2.5-7B, standard, CrewAI & \cellcolor[HTML]{7C7CFF} 11.7 (3.2) & \cellcolor[HTML]{ADADFF} 830.6 (168.6) \\
Qwen2.5-7B, MCP, CrewAI & \cellcolor[HTML]{7373FF} 10.3 (3.7) & \cellcolor[HTML]{8282FF} 764.7 (60.1) \\
Qwen2.5-7B, A2A, CrewAI & \cellcolor[HTML]{7979FF} 11.2 (6.7) & \cellcolor[HTML]{9F9FFF} 809.9 (73.7) \\ \hline

Qwen2.5-7B, standard, LangChain & \cellcolor[HTML]{8484FF} 13.1 (2.8) & \cellcolor[HTML]{9B9BFF} 802.4 (167.5) \\
Qwen2.5-7B, MCP, LangChain & \cellcolor[HTML]{6F6FFF} \textbf{9.6} (3.0) & \cellcolor[HTML]{7878FF} 749.3 (71.4) \\
Qwen2.5-7B, A2A, LangChain & \cellcolor[HTML]{7C7CFF} 11.7 (2.8) & \cellcolor[HTML]{9696FF} 795.2 (83.6) \\ \hline

Qwen2.5-7B, standard, LangGraph & \cellcolor[HTML]{8686FF} 13.4 (2.8) & \cellcolor[HTML]{9C9CFF} 804.5 (170.2) \\
Qwen2.5-7B, MCP, LangGraph & \cellcolor[HTML]{7070FF} 9.7 (6.7) & \cellcolor[HTML]{7373FF} \textbf{741.8} (\textbf{\textit{57.3}}) \\
Qwen2.5-7B, A2A, LangGraph & \cellcolor[HTML]{7979FF} 11.3 (2.9) & \cellcolor[HTML]{9B9BFF} 802.4 (87.2) \\ \hline

Qwen2.5-7B, standard, Semantic Kernel & \cellcolor[HTML]{8383FF} 13.0 (\textbf{\textit{2.0}}) & \cellcolor[HTML]{A6A6FF} 819.6 (173.9) \\
Qwen2.5-7B, MCP, Semantic Kernel & \cellcolor[HTML]{7676FF} 10.8 (11.5) & \cellcolor[HTML]{7878FF} 749.6 (69.0) \\
Qwen2.5-7B, A2A, Semantic Kernel & \cellcolor[HTML]{7979FF} 11.2 (2.8) & \cellcolor[HTML]{9A9AFF} 802.1 (83.5) \\ \hline

Llama-4-Scout-17B-16E, standard, CrewAI & \cellcolor[HTML]{9393FF} 15.7 (15.6) & \cellcolor[HTML]{7D7DFF} 756.9 (137.4) \\
Llama-4-Scout-17B-16E, MCP, CrewAI & \cellcolor[HTML]{6969FF} \textbf{8.6} (\textbf{3.8}) & \cellcolor[HTML]{BEBEFF} 856.5 (95.3) \\
Llama-4-Scout-17B-16E, A2A, CrewAI & \cellcolor[HTML]{7979FF} 11.2 (7.1) & \cellcolor[HTML]{FFFEFE} 961.7 (111.7) \\ \hline

Llama-4-Scout-17B-16E, standard, LangChain & \cellcolor[HTML]{7575FF} 10.6 (7.7) & \cellcolor[HTML]{6969FF} \textbf{726.5} (137.0) \\
Llama-4-Scout-17B-16E, MCP, LangChain & \cellcolor[HTML]{7070FF} 9.7 (5.6) & \cellcolor[HTML]{AFAFFF} 834.1 (91.6) \\
Llama-4-Scout-17B-16E, A2A, LangChain & \cellcolor[HTML]{8282FF} 12.7 (7.6) & \cellcolor[HTML]{F7F7FF} 944.4 (117.8) \\ \hline

Llama-4-Scout-17B-16E, standard, LangGraph & \cellcolor[HTML]{7373FF} 10.3 (4.9) & \cellcolor[HTML]{6F6FFF} 735.9 (139.5) \\
Llama-4-Scout-17B-16E, MCP, LangGraph & \cellcolor[HTML]{7070FF} 9.8 (6.5) & \cellcolor[HTML]{ADADFF} 831.2 (\textbf{82.0}) \\
Llama-4-Scout-17B-16E, A2A, LangGraph & \cellcolor[HTML]{7A7AFF} 11.5 (4.8) & \cellcolor[HTML]{F4F4FF} 939.5 (114.1) \\ \hline

Llama-4-Scout-17B-16E, standard, Semantic Kernel & \cellcolor[HTML]{FFE1E1} 45.8 (25.5) & \cellcolor[HTML]{A1A1FF} 812.9 (198.6) \\
Llama-4-Scout-17B-16E, MCP, Semantic Kernel & \cellcolor[HTML]{FFD2D2} 51.5 (27.8) & \cellcolor[HTML]{C0C0FF} 860.1 (143.8) \\
Llama-4-Scout-17B-16E, A2A, Semantic Kernel & \cellcolor[HTML]{FFC0C0} 58.8 (30.6) & \cellcolor[HTML]{E7E7FF} 919.1 (189.9) \\

\bottomrule
\end{tabular}
\end{table*}

\clearpage

\subsection{Tool Use}

Tool use was computed according to Equation \ref{eq:tool_use}. In the results that follow, the only model that externalized the use of tools in this experiment was Qwen2.5-7B. Hence, the proper analysis combining tool usage by all combinations was affected by model behaviors of answering using internal knowledge or not externalizing the tool use.


Kruskal-Wallis test for tool use rate suggests that there are significant differences for protocol (H-statistic = 11.636, $p-value$ = 0.003) and framework (H-statistic = 32.818, $p-value$ < 0.001).

Dunn's Post-hoc suggests significant pairwise differences for protocol, when assessing tool usage, as follows: 
\begin{itemize}
\item A2A vs. MCP ($p-value$ = 0.002).
\end{itemize}

Dunn's Post-hoc suggests significant pairwise differences for framework, when assessing tool usage, as follows:
\begin{itemize}
\item CrewAI vs. LangChain ($p-value$ < 0.001),
\item CrewAI vs. LangGraph ($p-value$ < 0.001), and
\item CrewAI vs. SemanticKernel ($p-value$ < 0.001).
\end{itemize}

\begin{figure*}[h]
  \centering
  \includegraphics[width=0.8\linewidth,trim={21.5cm 0 0 0},clip]{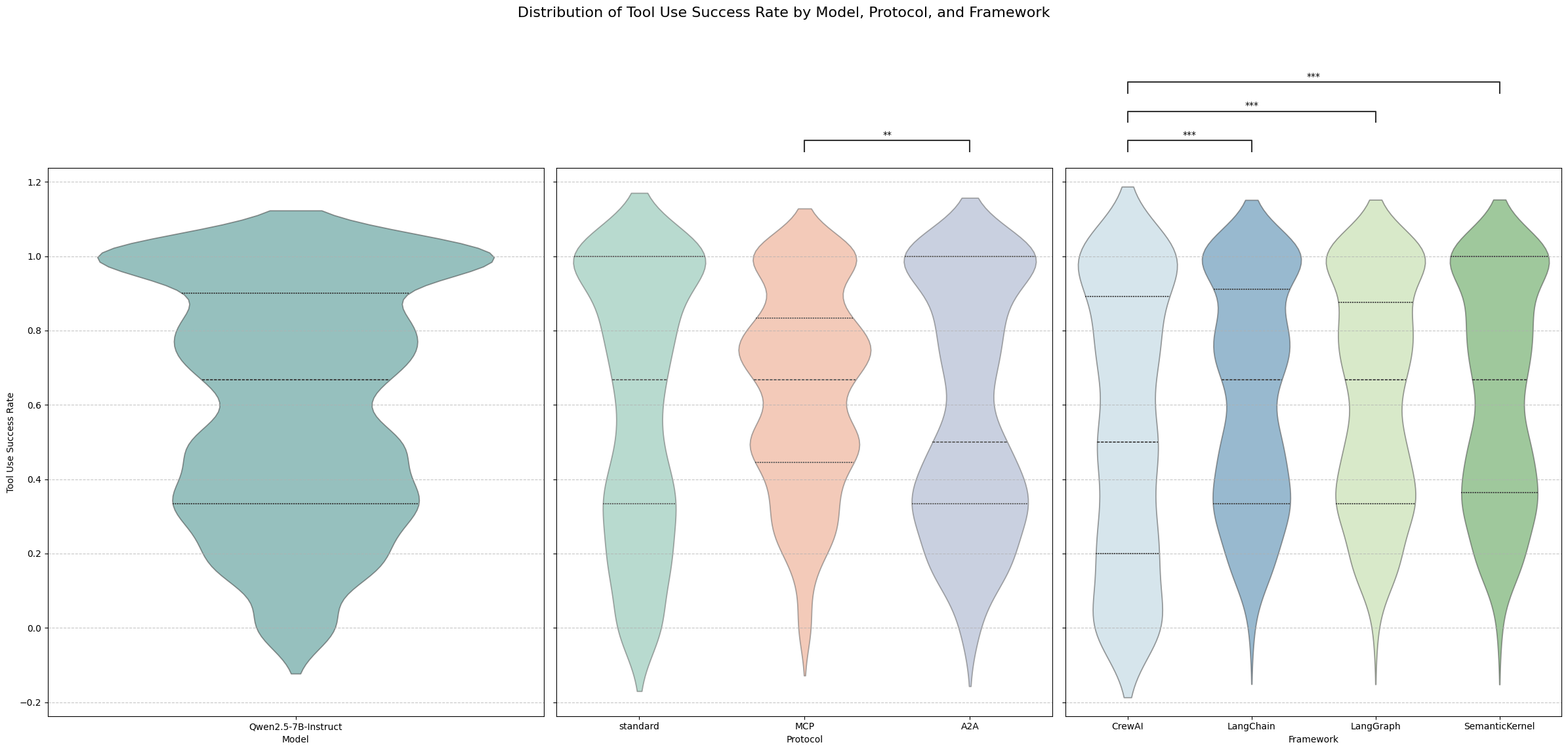}
  \caption{Tool use success rate for Qwen2.5-7B by protocol and framework. Dunn's Post-hoc pairwise significant differences are shown according to the following coding: *** for $p-value$ < 0.001, ** for $p-value$ < 0.01, and * for $p-value$ < 0.05.}
  \Description{Tool use success rate by model, protocol, and framework.}
  \label{fig:tool_use}
\end{figure*}

Figure \ref{fig:tool_use} shows a violin plot grouping tool use success by protocol and framework.
Considering significant differences found and Kruskal-Wallis mean ranks per group, results suggest that \textbf{MCP} (mean rank=1856.70), and
\textbf{SemanticKernel} (mean rank=1856.14) are a combination that promoted tool use and obtained the best results under the tested conditions.

\begin{table*}
\caption{Tool use success rate for different technology stacks. Not shown technology stacks did not register tool usage.}
\label{tab:tool_use}
\begin{tabular}{lr}
\toprule

$TS_{model, protocol, framework}$ & $TUSR_{TS}$ ($\sigma$) \\

\midrule

Qwen2.5-7B, standard, CrewAI & \cellcolor[HTML]{FFD0D0} 0.543 (0.442) \\
Qwen2.5-7B, MCP, CrewAI & \cellcolor[HTML]{FFC0C0} 0.523 (0.306) \\
Qwen2.5-7B, A2A, CrewAI & \cellcolor[HTML]{FFE6E6} 0.570 (0.330) \\ \hline

Qwen2.5-7B, standard, LangChain & \cellcolor[HTML]{BBBBFF} 0.638 (0.310) \\
Qwen2.5-7B, MCP, LangChain & \cellcolor[HTML]{7E7EFF} 0.670 (0.238) \\
Qwen2.5-7B, A2A, LangChain & \cellcolor[HTML]{FFEBEB} 0.577 (0.323) \\ \hline

Qwen2.5-7B, standard, LangGraph & \cellcolor[HTML]{C6C6FF} 0.632 (0.313)\\
Qwen2.5-7B, MCP, LangGraph & \cellcolor[HTML]{8787FF} 0.665 (\textbf{\textit{0.229}}) \\
Qwen2.5-7B, A2A, LangGraph & \cellcolor[HTML]{FFF2F2} 0.586 (0.327) \\ \hline

Qwen2.5-7B, standard, Semantic Kernel & \cellcolor[HTML]{F2F2FF} 0.609 (0.317) \\
Qwen2.5-7B, MCP, Semantic Kernel & \cellcolor[HTML]{6969FF} \textbf{0.681} (0.243) \\
Qwen2.5-7B, A2A, Semantic Kernel & \cellcolor[HTML]{FDFDFF} 0.603 (0.317) \\

\bottomrule
\end{tabular}
\end{table*}

\subsection{Social Values}

Social value stability score was computed according to Equation \ref{eq:stack_value_stability}. 
Kruskal-Wallis test for social value scores suggests that there are significant differences for model (H-statistic = 1081.219, $p-value$ < 0.001), protocol (H-statistic = 6.534, $p-value$ = 0.038), and framework (H-statistic = 73.181, $p-value$ < 0.001).

Dunn's Post-hoc suggests significant pairwise differences for model, when assessing social value stability, as follows:
\begin{itemize}
\item Llama-4-Scout-17B-16E vs. Mistral-7B ($p-value$ < 0.001) and
\item Llama-4-Scout-17B-16E vs. Qwen2.5-7B ($p-value$ < 0.001).
\end{itemize}

Dunn's Post-hoc suggests significant pairwise differences for protocol, when assessing social value stability, as follows:
\begin{itemize}
\item A2A vs. standard ($p-value$ = 0.032).
\end{itemize}

Dunn's Post-hoc suggests significant pairwise differences for framework, when assessing social value stability, as follows:
\begin{itemize}
\item CrewAI vs. SemanticKernel ($p-value$ < 0.001),
\item LangChain vs. SemanticKernel ($p-value$ < 0.001), and
\item LangGraph vs. SemanticKernel ($p-value$ = 0.025).
\end{itemize}

\begin{figure*}[h]
  \centering
  \includegraphics[width=\linewidth]{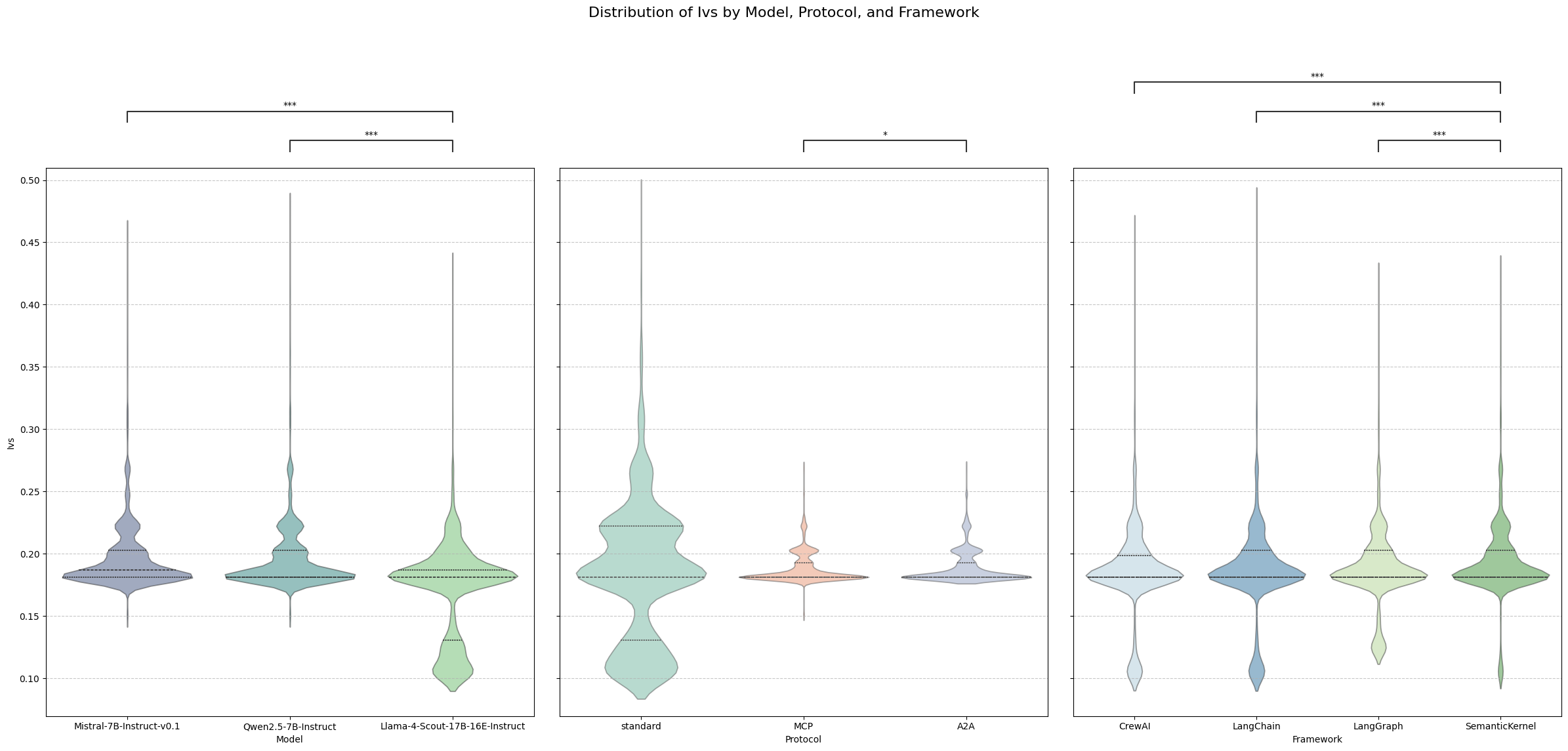}
  \caption{Social value score by model, protocol, and framework. Dunn's Post-hoc pairwise significant differences are shown according to the following coding: *** for $p-value$ < 0.001, ** for $p-value$ < 0.01, and * for $p-value$ < 0.05.}
  \Description{Social value score by model, protocol, and framework.}
  \label{fig:social_values}
\end{figure*}

Figure \ref{fig:social_values} shows a violin plot grouping social value scores by model, protocol, and framework.
Considering significant differences found and Kruskal-Wallis mean ranks per group, results suggest that the best-performing technology stack in terms of social value stability should combine:
\textbf{Mistral-7B} (mean rank=6100.76),
\textbf{A2A} (mean rank=5460.16), and
\textbf{SemanticKernel} (mean rank=5778.08).

\newpage

Table \ref{tab:social_value} shows results obtained in terms of social value stability for input task to output, reasoning steps, and social value stability throughout all the steps, from input task up to final answer.
It is possible to verify that social values related to input task and output (final answer) are less stable than reasoning steps. Reasoning steps are more stable in the whole process, often representing the same values in its intermediate reasoning steps.

\begin{figure}[htbp]
    \centering

    \begin{subfigure}[b]{0.45\textwidth}
        \centering
        \includegraphics[width=\textwidth]{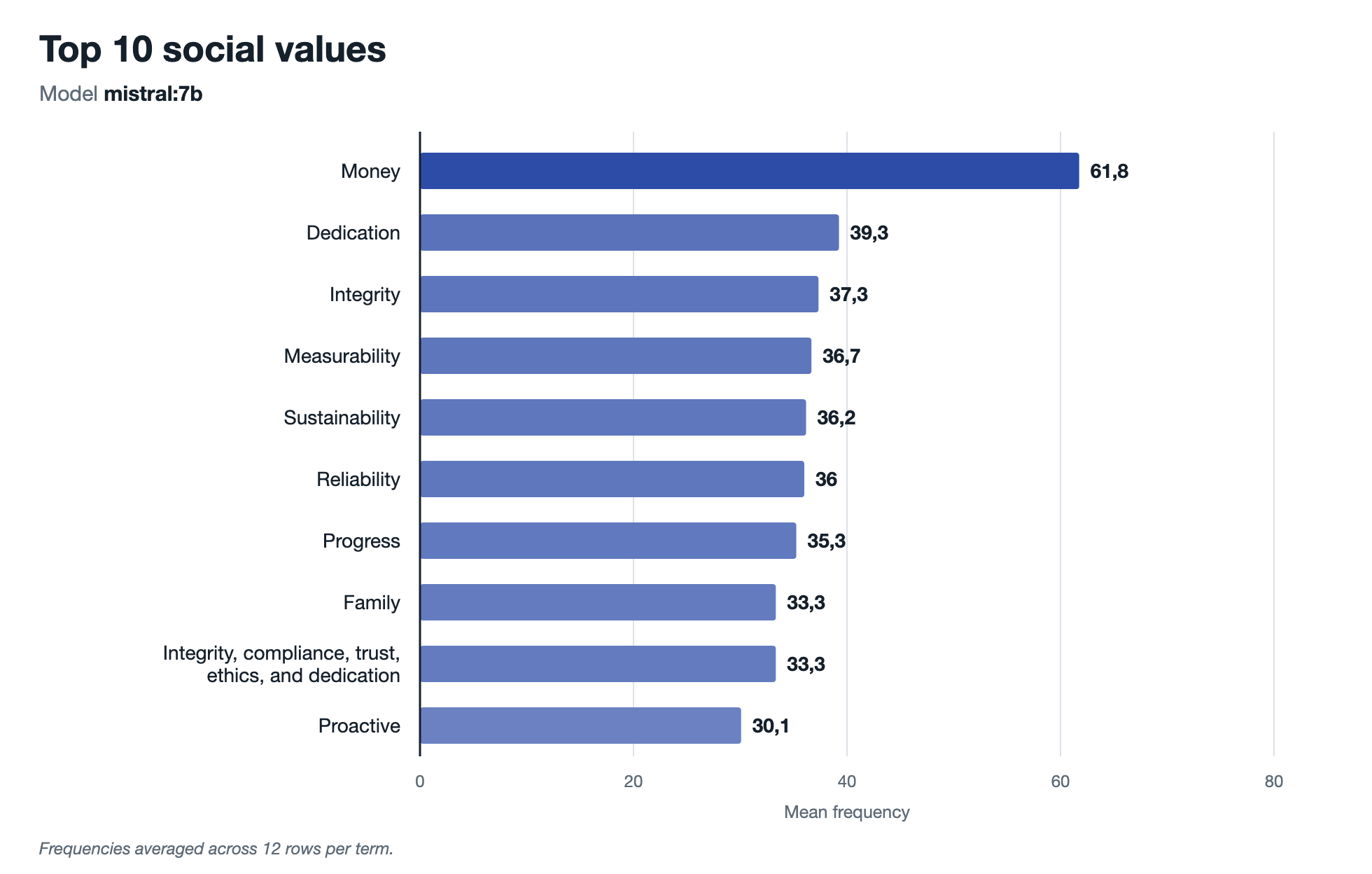}
        \caption{Top 10 social values for input tasks\newline provided to Mistral-7B-Instruct-v0.1.}
        \label{fig:top_values_mistral_input}
    \end{subfigure}
    \hfill
    \begin{subfigure}[b]{0.45\textwidth}
        \centering
        \includegraphics[width=\textwidth]{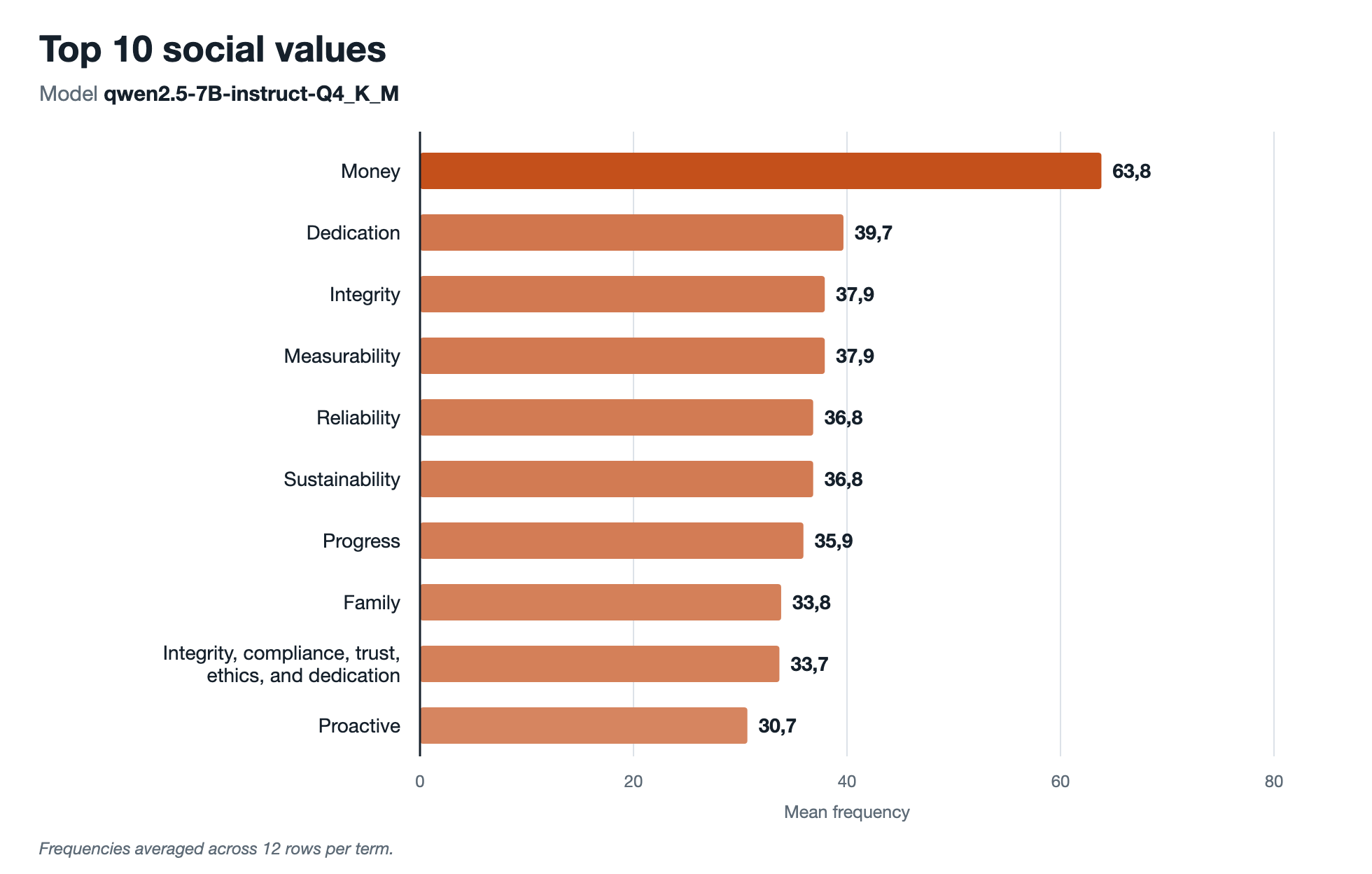}
        \caption{Top 10 social values for input tasks\newline provided to Qwen2.5-7B.}
        \label{fig:top_values_qwen_input}
    \end{subfigure}

    \vspace{1em} 
    
    \begin{subfigure}[b]{0.45\textwidth}
        \centering
        \includegraphics[width=\textwidth]{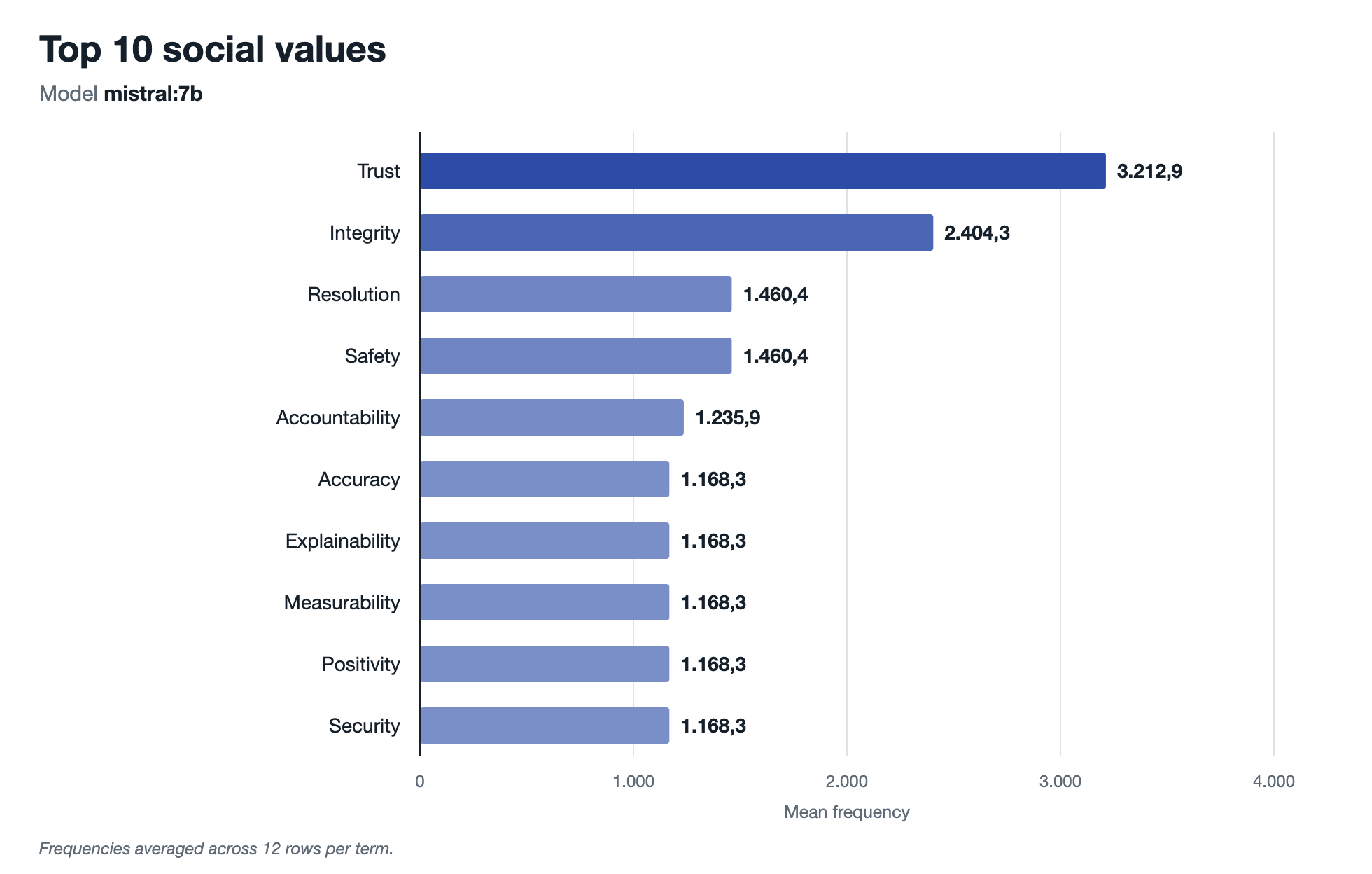}
        \caption{Top 10 social values for reasoning steps\newline using Mistral-7B-Instruct-v0.1.}
        \label{fig:top_values_mistral_reasoning}
    \end{subfigure}
    \hfill
    \begin{subfigure}[b]{0.45\textwidth}
        \centering
        \includegraphics[width=\textwidth]{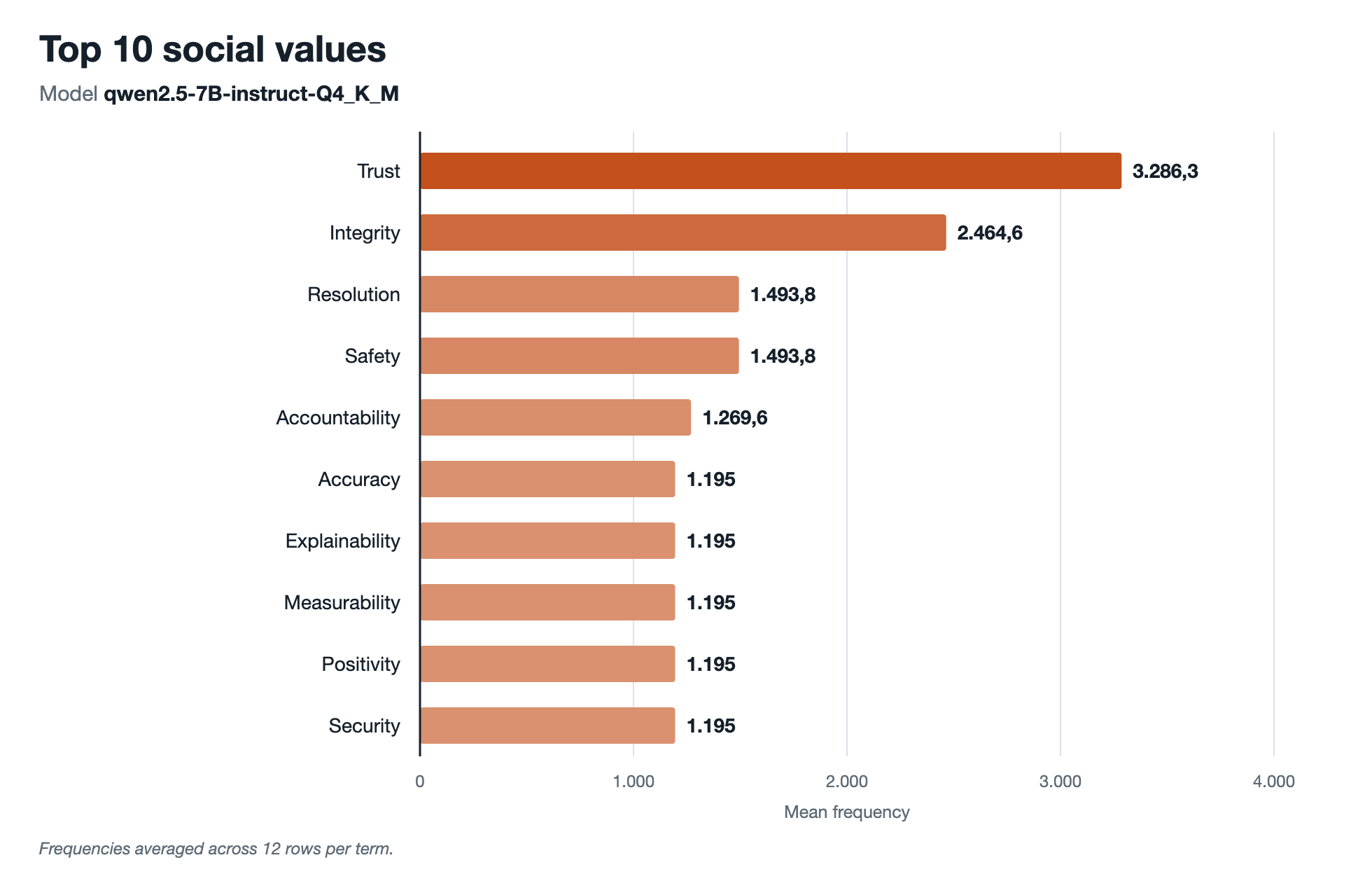}
        \caption{Top 10 social values for reasoning steps\newline using Qwen2.5-7B.}
        \label{fig:top_values_qwen_reasoning}
    \end{subfigure}

    \vspace{1em} 
    
    \begin{subfigure}[b]{0.45\textwidth}
        \centering
        \includegraphics[width=\textwidth]{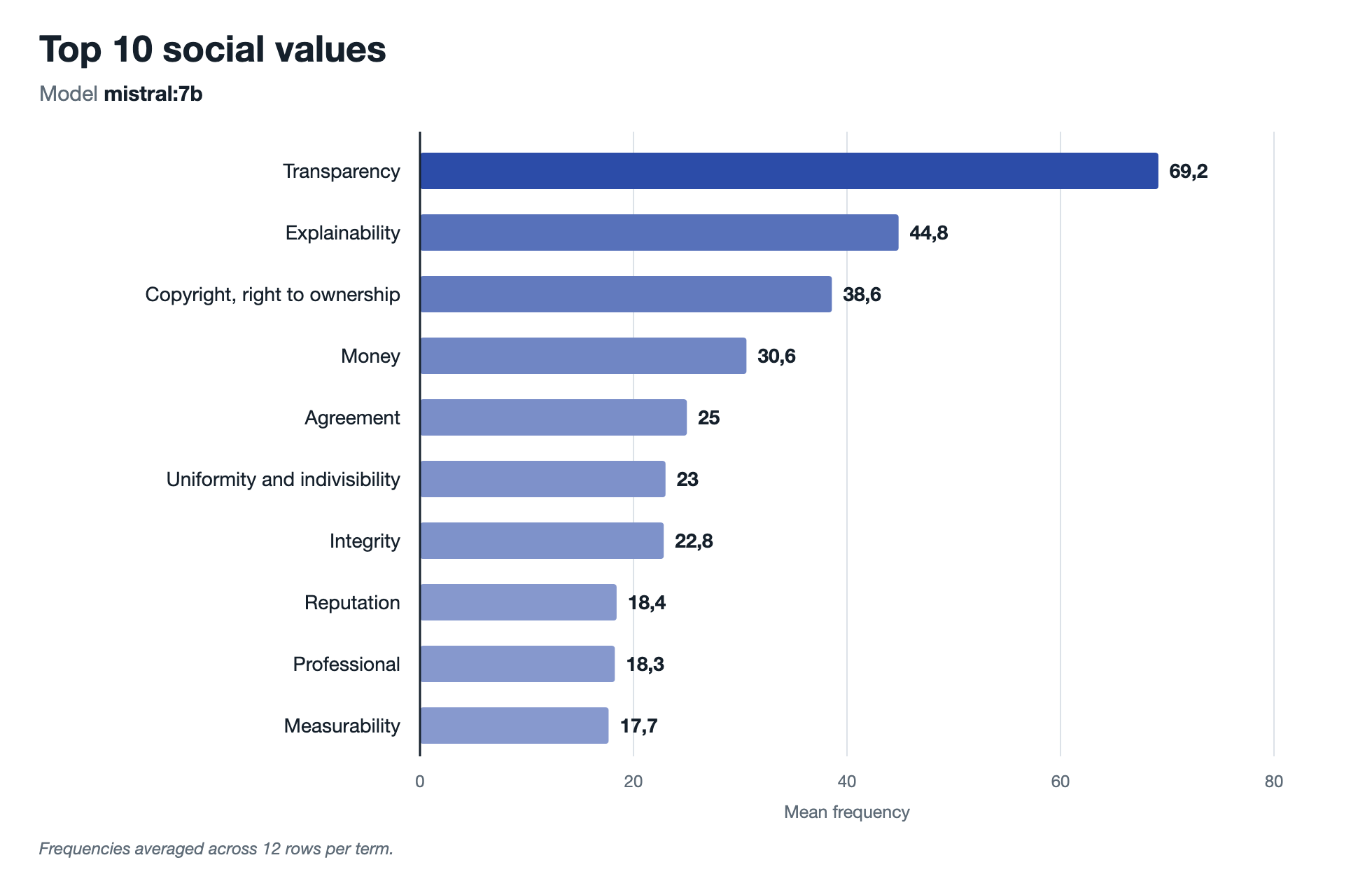}
        \caption{Top 10 social values for final output\newline provided using Mistral-7B-Instruct-v0.1.}
        \label{fig:top_values_mistral_output}
    \end{subfigure}
    \begin{subfigure}[b]{0.45\textwidth}
        \centering
        \includegraphics[width=\textwidth]{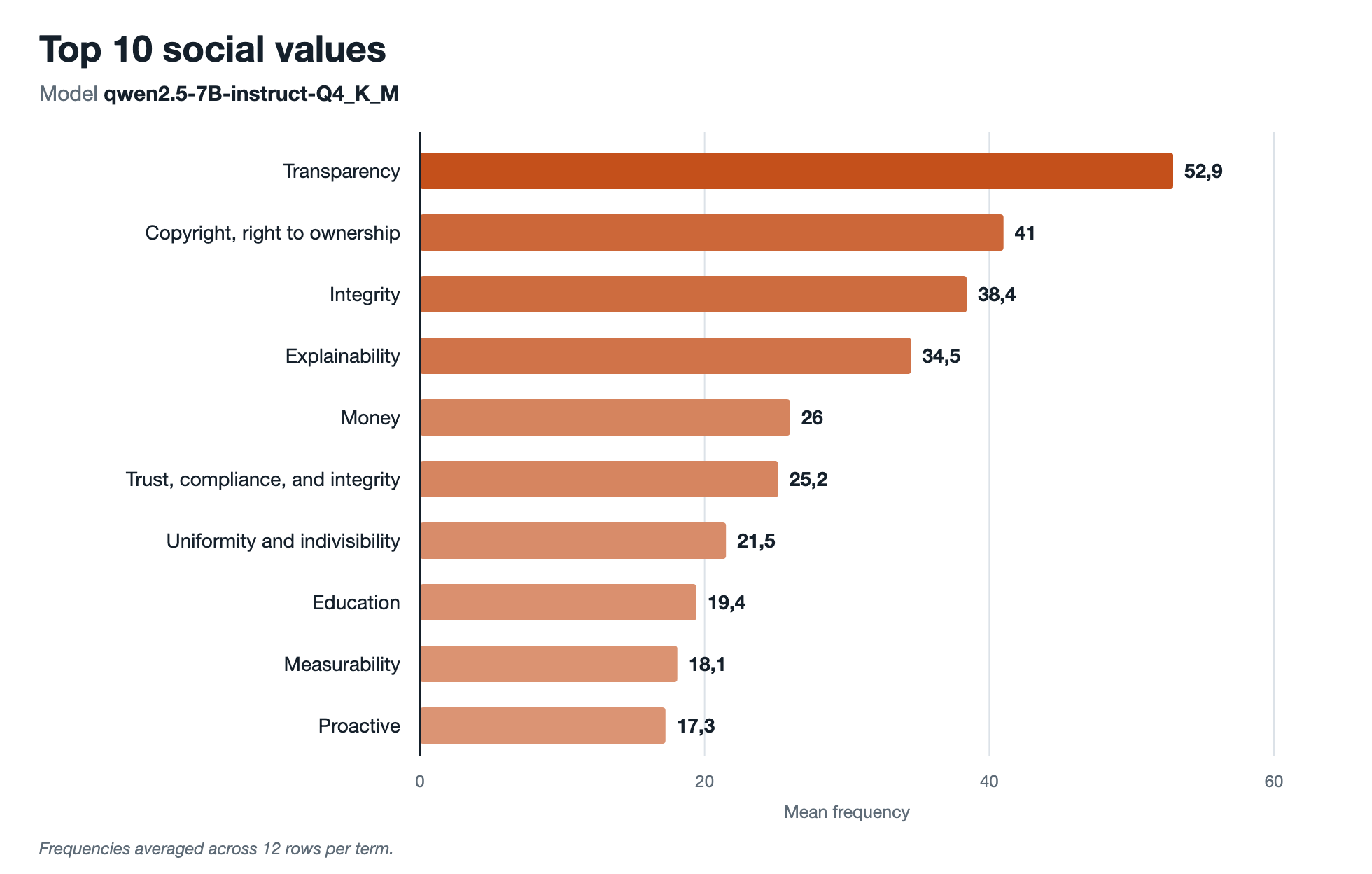}
        \caption{Top 10 social values for final output\newline provided using Qwen2.5-7B.}
        \label{fig:top_values_qwen_output}
    \end{subfigure}
    
    \caption{Comparison of mean frequency of social values retrieved by a human-curated prompt social value classifier \cite{Santana2025}. Input tasks (a, b) show that classifier is stable. Reasoning steps (c, d) show that intermediate reasoning steps are consistent reporting agents functioning, regardless of input. Social values represented in the final answer (e, f) are similar, differing in the ranking involving some values.}
    \label{fig:top_values}
\end{figure}

Figure \ref{fig:top_values} shows the most common social values present in input tasks, reasoning steps, and final answer output.
In the figure it is possible to verify that Mind2Web tasks influence the values identified. In reasoning steps, prompts exchanged between agents are stable in terms of value and provide results and content for the next step, highlighted by values such as trust, integrity, safety, and accountability. Social values present in the outputs are also somewhat consistent between models. Transparency, explainability, and right to ownership are present in the top 5 most frequent values for both 7B models shown.

\begin{table*}
\caption{Social value stability considering values present in the prompt and final answer ($SVS_{{TS}_{P,o_a}}$), reasoning steps ($SVS_{{TS}_{reasoning}}$), and all steps from prompt, reasoning steps up to final answer ($SVS_{TS}$) for different technology stacks; the higher the better.}
\label{tab:social_value}
\begin{tabular}{lrrr}
\toprule

$TS_{model, protocol, framework}$ & $SVS_{{TS}_{P,o_a}}$ & $SVS_{{TS}_{reasoning}}$ & $SVS_{TS}$ ($\sigma$) \\

\midrule

Mistral-7B, standard, CrewAI & \cellcolor[HTML]{EEEEFF} 0.156 & \cellcolor[HTML]{6E6EFF} 0.235 & \cellcolor[HTML]{9393FF} 0.212 (0.0380) \\
Mistral-7B, MCP, CrewAI & \cellcolor[HTML]{FFF4F4} 0.129 & \cellcolor[HTML]{7979FF} 0.228 & \cellcolor[HTML]{B8B8FF} 0.189 (\textbf{\textit{0.0116}}) \\ 
Mistral-7B, A2A, CrewAI & \cellcolor[HTML]{FFE3E3} 0.105 & \cellcolor[HTML]{7979FF} 0.228 & \cellcolor[HTML]{B8B8FF} 0.189 (0.0123) \\ \hline

Mistral-7B, standard, LangChain & \cellcolor[HTML]{FEFEFF} 0.146 & \cellcolor[HTML]{6E6EFF} 0.235 & \cellcolor[HTML]{9393FF} 0.212 (0.0383) \\
Mistral-7B, MCP, LangChain & \cellcolor[HTML]{FFDADA} 0.091 & \cellcolor[HTML]{7979FF} 0.228 & \cellcolor[HTML]{B8B8FF} 0.189 (0.0124) \\ 
Mistral-7B, A2A, LangChain & \cellcolor[HTML]{D4D4FF} \textbf{0.172} & \cellcolor[HTML]{7979FF} 0.228 & \cellcolor[HTML]{B7B7FF} 0.190 (0.0134) \\ \hline

Mistral-7B, standard, LangGraph & \cellcolor[HTML]{FFFFFF} 0.145 & \cellcolor[HTML]{6B6BFF} \textbf{0.237} & \cellcolor[HTML]{9292FF} \textbf{0.213} (0.0385) \\
Mistral-7B, MCP, LangGraph & \cellcolor[HTML]{FFF0F0} 0.123 & \cellcolor[HTML]{7979FF} 0.228 & \cellcolor[HTML]{B7B7FF} 0.190 (0.0126) \\ 
Mistral-7B, A2A, LangGraph & \cellcolor[HTML]{FFE6E6} 0.108 & \cellcolor[HTML]{7979FF} 0.228 & \cellcolor[HTML]{B7B7FF} 0.190 (0.0126) \\ \hline

Mistral-7B, standard, Semantic Kernel & \cellcolor[HTML]{FFFDFD} 0.143 & \cellcolor[HTML]{6969FF} 0.238 & \cellcolor[HTML]{9292FF} \textbf{0.213} (0.0395) \\
Mistral-7B, MCP, Semantic Kernel & \cellcolor[HTML]{FFE7E7} 0.110 & \cellcolor[HTML]{7979FF} 0.228 & \cellcolor[HTML]{B8B8FF} 0.189 (0.0115) \\ 
Mistral-7B, A2A, Semantic Kernel & \cellcolor[HTML]{FFE9E9} 0.113 & \cellcolor[HTML]{7979FF} 0.228 & \cellcolor[HTML]{B7B7FF} 0.190 (0.0141) \\ \hline

Qwen2.5-7B, standard, CrewAI & \cellcolor[HTML]{FFF7F7} 0.134 & \cellcolor[HTML]{6E6EFF} 0.235 & \cellcolor[HTML]{9696FF} 0.210 (\textbf{\textit{0.0368}}) \\
Qwen2.5-7B, MCP, CrewAI & \cellcolor[HTML]{FFE5E5} 0.107 & \cellcolor[HTML]{7979FF} 0.228 & \cellcolor[HTML]{BABAFF} 0.188 (0.0114) \\
Qwen2.5-7B, A2A, CrewAI & \cellcolor[HTML]{FFFFFF} 0.145 & \cellcolor[HTML]{7979FF} 0.228 & \cellcolor[HTML]{B7B7FF} 0.190 (0.0139) \\ \hline

Qwen2.5-7B, standard, LangChain & \cellcolor[HTML]{CFCFFF} \textbf{0.175} & \cellcolor[HTML]{6E6EFF} 0.235 & \cellcolor[HTML]{9595FF} 0.211 (0.0379) \\
Qwen2.5-7B, MCP, LangChain & \cellcolor[HTML]{FFD6D6} 0.085 & \cellcolor[HTML]{7979FF} 0.228 & \cellcolor[HTML]{BABAFF} 0.188 (0.0121) \\
Qwen2.5-7B, A2A, LangChain & \cellcolor[HTML]{FFFEFE} 0.144 & \cellcolor[HTML]{7979FF} 0.228 & \cellcolor[HTML]{B7B7FF} 0.190 (0.0135) \\ \hline

Qwen2.5-7B, standard, LangGraph & \cellcolor[HTML]{FFF0F0} 0.124 & \cellcolor[HTML]{6B6BFF} \textbf{0.237} & \cellcolor[HTML]{9595FF} 0.211 (0.0370) \\
Qwen2.5-7B, MCP, LangGraph & \cellcolor[HTML]{FFD4D4} 0.082 & \cellcolor[HTML]{7979FF} 0.228 & \cellcolor[HTML]{BABAFF} 0.188 (0.0125) \\
Qwen2.5-7B, A2A, LangGraph & \cellcolor[HTML]{FFECEC} 0.117 & \cellcolor[HTML]{7979FF} 0.228 & \cellcolor[HTML]{B7B7FF} 0.190 (0.0135) \\ \hline

Qwen2.5-7B, standard, Semantic Kernel & \cellcolor[HTML]{FFFCFC} 0.141 & \cellcolor[HTML]{6E6EFF} 0.235 & \cellcolor[HTML]{9393FF} \textbf{0.212} (0.0376) \\
Qwen2.5-7B, MCP, Semantic Kernel & \cellcolor[HTML]{FFF1F1} 0.125 & \cellcolor[HTML]{7979FF} 0.228 & \cellcolor[HTML]{BABAFF} 0.188 (0.0126) \\
Qwen2.5-7B, A2A, Semantic Kernel & \cellcolor[HTML]{F4F4FF} 0.152 & \cellcolor[HTML]{7979FF} 0.228 & \cellcolor[HTML]{B8B8FF} 0.189 (0.0128) \\ \hline

Llama-4-Scout-17B, standard, CrewAI & \cellcolor[HTML]{FFF2F2} 0.126 & \cellcolor[HTML]{FFF6F6} 0.133 & \cellcolor[HTML]{FFE7E7} 0.110 (0.0101) \\
Llama-4-Scout-17B, MCP, CrewAI & \cellcolor[HTML]{FFD7D7} 0.086 & \cellcolor[HTML]{7979FF} 0.228 & \cellcolor[HTML]{BABAFF} 0.188 (0.0128) \\
Llama-4-Scout-17B, A2A, CrewAI & \cellcolor[HTML]{FFD7D7} 0.086 & \cellcolor[HTML]{7979FF} 0.228 & \cellcolor[HTML]{B8B8FF} 0.189 0.0132) \\ \hline

Llama-4-Scout-17B, standard, LangChain & \cellcolor[HTML]{FFD5D5} 0.084 & \cellcolor[HTML]{FFF6F6} 0.133 & \cellcolor[HTML]{FFE7E7} 0.110 (\textbf{0.0094}) \\
Llama-4-Scout-17B, MCP, LangChain & \cellcolor[HTML]{FFDFDF} 0.098 & \cellcolor[HTML]{7979FF} 0.228 & \cellcolor[HTML]{BABAFF} 0.188 (0.0126) \\
Llama-4-Scout-17B, A2A, LangChain & \cellcolor[HTML]{FFD7D7} 0.087 & \cellcolor[HTML]{7979FF} 0.228 & \cellcolor[HTML]{BABAFF} 0.188 (0.0131) \\ \hline

Llama-4-Scout-17B, standard, LangGraph & \cellcolor[HTML]{FFDFDF} 0.098 & \cellcolor[HTML]{EBEBFF} 0.158 & \cellcolor[HTML]{FFF4F4} 0.130 (0.0105) \\
Llama-4-Scout-17B, MCP, LangGraph & \cellcolor[HTML]{FFD4D4} 0.083 & \cellcolor[HTML]{7979FF} 0.228 & \cellcolor[HTML]{B8B8FF} 0.189 (0.0137) \\
Llama-4-Scout-17B, A2A, LangGraph & \cellcolor[HTML]{FFC0C0} 0.053 & \cellcolor[HTML]{7979FF} 0.228 & \cellcolor[HTML]{B7B7FF} \textbf{0.190} (0.0143) \\ \hline

Llama-4-Scout-17B, standard, Semantic Kernel & \cellcolor[HTML]{9595FF} \textbf{0.211} & \cellcolor[HTML]{BCBCFF} 0.187 & \cellcolor[HTML]{BCBCFF} 0.187 (0.0551) \\
Llama-4-Scout-17B, MCP, Semantic Kernel & \cellcolor[HTML]{FFEEEE} 0.120 & \cellcolor[HTML]{7979FF} 0.228 & \cellcolor[HTML]{BABAFF} 0.188 (0.0123) \\
Llama-4-Scout-17B, A2A, Semantic Kernel & \cellcolor[HTML]{ECECFF} 0.157 & \cellcolor[HTML]{7979FF} 0.228 & \cellcolor[HTML]{B8B8FF} 0.189 (0.0121) \\

\bottomrule
\end{tabular}
\end{table*}

\clearpage

\section{Discussion}
\label{sec:discussion}

In this section we discuss outcomes obtained according to each of the metrics studied defined in Section \ref{sec:problem_definition}.
Table \ref{tab:best_stack_summary} provides an overview of Agentic Web technology stacks results and associated research questions (RQs).

\begin{table}[h]
  \small
  \caption{Best-performing technology stacks per metric (consolidated from Sections 5.1 to 5.4) and respective research questions (from RQ1 to RQ4)}
  \label{tab:best_stack_summary}
  \begin{tabular}{rllll}
    \toprule
    & \textbf{Metric} & \textbf{Model} & \textbf{Protocol} & \textbf{Framework}\\
    \midrule
    \multirow{3}{2.5cm}{Efficiency (RQ2) $~\Biggl\{$} & Task Completion (RQ1) & Qwen2.5-7B & Standard & SemanticKernel \\
    & Convergence Time & Qwen2.5-7B & MCP & LangChain \\
    & Verbosity & Qwen2.5-7B & Standard & LangChain  \\
    & Tool Use Success (RQ3) & Qwen2.5-7B\textsuperscript{*} & MCP & SemanticKernel \\
    & Social Value Stability (RQ4) & Mistral-7B & A2A & SemanticKernel \\
    \bottomrule
  \end{tabular}
  \\ \footnotesize{\textsuperscript{*}Only model that externalized tool calls.}
\end{table}

\subsection{Task Completion}

Bearing mind of RQ1 on task completion rate under the conditions studied, results suggest that \textbf{Qwen2.5-7B}, \textbf{LangChain} and \textbf{standard} conditions represent the best-performing baseline.
\textbf{Qwen2.5-7B} with other conditions, considering protocol messages MCP and A2A, obtained worse results. However, when using standard messages, it obtained better results. Beyond task completion, Table \ref{tab:tcr} also shows that models often understood the proposed task and that reasoning steps kept track of the task at hand. Lower values for task understanding and task adherence occurred when combining \textbf{Llama-4-Scout-17B-16E} with \textbf{MCP} and \textbf{A2A}, suggesting that testing how agents understand the task and whether or not internal reasoning steps continue on track has an interesting value in such evaluations.
Finally, Table \ref{tab:tcr} also reveals a pattern in which task completion is impacted under \textbf{Qwen2.5-7B} and \textbf{MCP} conditions.

\subsection{Efficiency}

Bearing mind of RQ2 on efficiency, which combines task completion rate, convergence time, and verbosity, results suggest that \textbf{Qwen2.5-7B}, \textbf{LangChain} and \textbf{standard} conditions represent the best-performing balance combining task completion, low convergence time, and low verbosity (Table \ref{tab:best_stack_summary}).

Figure \ref{fig:efficiency_pareto_frontier} presented a Pareto frontier assessment allowing the identification of different nuances in the combination of task completion rates, convergence time, and verbosity.
Better task completion scores were obtained under 
\textbf{Qwen2.5-7B, standard, LangChain} and 
\textbf{Qwen2.5-7B, standard, Semantic Kernel} conditions
Better balance on low convergence time and low verbosity were obtained under 
\textbf{Qwen2.5-7B, A2A, Semantic Kernel}
\textbf{Qwen2.5-7B, A2A, LangChain} conditions
Finally, technology stacks with low convergence time and average
task completion were obtained under
\textbf{Mistral-7B, standard, LangChain}
\textbf{Mistral-7B, standard, LangGraph} conditions.

Finally, Table \ref{tab:best_stack_summary} also reveals an outlier under conditions \textbf{Llama-4-Scout-17B-16E} and \textbf{SemanticKernel}, given that they take more time to converge, but, at the same time, are not as verbose as other conditions.

\newpage

\subsection{Tool Use}

Bearing mind of RQ3 on tool use, results suggest that 
\textbf{Qwen2.5-7B}, \textbf{Semantic Kernel} and \textbf{MCP} conditions used more the provided tools.
Table \ref{tab:tool_use} and Figure \ref{fig:tool_use} also show that \textbf{MCP} was the condition that obtained best-performing tool use success rates, suggesting that MCP messages foment agents to select and use tools more appropriately. However, as presented previously, tool usage was only performed (or externalized) by agents powered by \textbf{Qwen2.5-7B} model. Hence, results here are not related to the technology stack, but only under protocol and framework conditions within \textbf{Qwen2.5-7B}. More details on this is detailed in the limitations section (Section \ref{sec:limitations}).

\subsection{Social Value}

Bearing in mind RQ4 on social value stability, results suggest that \textbf{Mistral-7B}, \textbf{A2A}, and \textbf{SemanticKernel} conditions were more stable regarding how they process, reason, and answer a given task (Figure \ref{fig:social_values}, Figure \ref{fig:top_values}, and Table \ref{tab:social_value}).
Figure \ref{fig:social_values} shows that input tasks were properly tagged according to Mind2Web tasks, reasoning steps remain stable covering the functioning of the agentic workflow, and that outputs between models have intersecting values.

\section{Conclusion}
\label{sec:conclusion}

This paper presented an empirical assessment of the Agentic Web technology stack.
We have presented that when maintaining the tasks fixed and varying models, frameworks, and protocols, results may differ. 
In consonance with \citet{Ehtesham2025}, we have also have shown that no single Agentic Web technology stack suffices for all metrics.
Each context must be carefully considered, goal metrics defined, and, finally, tasks must be tested in a systematic way.
In this sense, MASTIF has shown to be a valuable tool for standardizing empirical studies for the Agentic Web.
In other words, we should consider AI as a normal technology \cite{Narayanan2025} and assess its building blocks for our specific contexts, tasks, and controlling for costs \cite{Kapoor2024}.

Regarding research questions defined, conditions using \textbf{Qwen2.5-7B} as model, \textbf{SemanticKernel} and \textbf{LangChain} as implementation frameworks represent interesting starting points for ones planning on testing MAS-based solutions for tasks similar to the ones present in the Mind2Web benchmark.

Tool calling assessment was impacted by how models unpack tasks and select tools to use. This is consonant with literature results that show that different models often differ on how they trigger tool use in MAS \cite{Santana2026}.

Beyond supporting empirical studies for the Agentic Web, this work also presents a proposal for assessing social value stability within agentic workflows.
While Fuse was proposed as a MAS simulation framework for user-mediated social reasoning \cite{Taubenfeld2026}, 
social value stability assessment provided by MASTIF is, to the best of our knowledge, a novel approach for assessing the entire flow from prompts, passing through reasoning steps, up to the final MAS answer.

In terms of social value stability, conditions using \textbf{Semantic Kernel} obtained the best-performing social value stability scores. Results represent a baseline for how MAS decompose task and how agents perform reasoning steps. The proposed score based on Jaccard Pairwise score supported the analysis of tasks, from input to output.

\newpage

\subsection{Lessons Learned}

In the development of MASTIF and running this experiment, a common issue was the fact that MAS systems count on multiple points of failure, with low observability, and highly non-deterministic behaviors. These characteristics make developing, running, debugging, and testing everything a Software Engineering `\textit{nightmare}.' 
Tool use is fragile. After troubleshooting and improving tools descriptions, so that models would identify when to use each tool, not all models used the tools provided.
In addition, given that tool call is \textit{non-deterministic}, it is impacted by model behavior, model size, embedded knowledge, temperature parameter, among others. In our initial tests, some models refuse to use tools informed, while others tried to use tools that don't exist (i.e., \textit{tool hallucination}). In terms of internal knowledge, larger models tend to provide answer using internal knowledge, while smaller models tend to use provided tools, as shown in the literature \cite{Santana2026}. Selecting models known to promote tool usage has shown to be of key importance when assessing MAS.

In terms of other factors of tool usage, it is important to provide MAS workflows descriptive information about available tools. In our first attempts, we tried using descriptive information about the tools, such as ``This tool does this, this, and that''. However, tools only started to be used in our experiments when we changed its description to guidance-like content, such as, ``Use this tool when you need to...''
Hence, tool's descriptions help models on identifying when using (or not) each tool. Hence, this required information impacts the input token burden, which may have an impact on token cost as well.

Regarding running tests using MASTIF, it already deals with multiple types of interruption, including quota errors, inference provider issues, etc. However, experiments using multiple models, protocols, and frameworks can result in huge log files detailing all reasoning and judge assessments. Hence, in experiment designs like the one detailed in this work, we suggest running each model in a separate step to allow easier assessment of log files and results.


\subsection{Reproducibility}

In terms of reproducibility of the experiment reported in this paper, the Mind2Web dataset is available on HuggingFace \footnote{https://huggingface.co/datasets/osunlp/Mind2Web}, the Multi-Agent System Testing Framework (MASTIF) developed is open source and available on GitHub \footnote{https://github.com/cewebbr/mastif/}, together with all parameters, models, prompts, and agentic frameworks considered detailed in this experiment's YAML file\footnote{\url{https://github.com/cewebbr/mastif/blob/main/experiments/tweb2026.yaml}}. Hence, we are providing all required information for others to reproduce and/or expand the experiment detailed in this paper.

\subsection{Ethics Statement}

This paper did not involve human participants and the benchmark of tasks considered is an open dataset, i.e., Mind2Web. Tasks from Mind2Web were sampled in a stratified way aiming at covering different tasks, different task complexities, different domains, and Web contexts.

\subsection{Limitations}
\label{sec:limitations}

The most notable limitations involve number of models used, judge, implementation frameworks, and tasks considered. All of these impact the cost to run such experiments \cite{Kapoor2024}. Given that all combinations must be tested, adding tasks, combining more models, and using setups with more than one judge --or more powerful ones-- result in a high number of task executions (Equation \ref{eq:tests}), increasing the cost for running such experiment from hundreds to thousands of dollars.
Moreover, the test involving different protocols considers the layer of how different protocols exchange messages. Testing with full implementations would require pre-built servers that provide all protocols studied, something that, to the best of our knowledge, is not available yet on the Web for services that would be representative for such study.

\subsection{Next Steps}

Results suggest that LLM-based agents and the Agentic Web technology stack should be assessed as any other technology \cite{Narayanan2025}, in an empirical and context-sensitive way, covering for multiple possibilities.
Hence, next steps of this research include studying the impact of different setups of MAS, including number of agents, hierarchical MAS, different sets of tools, max\_steps, temperature, agentic frameworks, new technologies such as WebMCP, and servers implementing MCP and A2A to support full protocol exchanges.

\begin{acks}
This work was supported by the Brazilian Network Information Center (NIC.br). 
Any opinions, findings, conclusions, or recommendations expressed in this material are those of the author(s) and do not necessarily reflect the views of NIC.br.
\end{acks}

\bibliographystyle{ACM-Reference-Format}
\bibliography{sample-base}

\appendix

\section{Agentic Frameworks}
\label{app:frameworks}

Table \ref{tab:framewors} lists LLM-agnostic agent frameworks in Python surveyed in this study.

\section{Benchmarks}
\label{app:benchmarks}

Table \ref{tab:benchmarks} lists benchmarks surveyed in this study that are related to LLM-based agents.

\section{Judge Model Rubric}
\label{app:judge_rubric}

Listing \ref{cod:judge_rubric} details the rubric used for judge model to assess information retrieved by the Agentic Web technology stacks assessed.

\begin{lstlisting}[caption={Rubric used by the judge model to assess responses to Mind2Web tasks.}, label={cod:judge_rubric}]
Evaluate the agent's response and reasoning on the following Mind2Web task.

Task:
{high_level_task}

Website: {website}
Domain: {domain}

Agent Response:
{agent_response}

Agent Reasoning:
{reasoning_steps_text or '(no reasoning provided)'}

For each of the following dimensions, provide a score from 0.0 to 1.0 and a brief rationale:

- task_understanding: how well the agent understood the task goals, constraints, and required output
- task_adherence: how well the agent's reasoning stayed on-task and relevant
- task_completion: whether the final response is sufficient to complete the task successfully

Use these anchors:

task_understanding:
1.0 = Explicitly captures all goals, constraints, and required outputs
0.7 = Captures main goal but misses minor constraints or details
0.5 = Understands general intent but misses key requirement(s)
0.3 = Misinterprets important parts of the task
0.0 = Completely incorrect or unrelated understanding

task_adherence:
1.0 = Fully focused on task, all steps directly relevant
0.7 = Mostly on-task, minor irrelevant or redundant steps
0.5 = Mixed relevance, some important off-track reasoning
0.3 = Frequently off-task or distracted reasoning
0.0 = Completely unrelated reasoning

task_completion:
1.0 = Fully completes task with correct and actionable steps/results
0.7 = Likely completes task but with minor gaps or inefficiencies
0.5 = Partially completes task; important steps missing
0.3 = Unlikely to complete task successfully
0.0 = Does not complete task at all or is incorrect

Respond with ONLY a JSON object with keys `task_understanding`, `task_adherence`, and `task_completion`.
Each value must be an object with keys `score` and `rationale`.
Example:
{{
  "task_understanding": {{"score": 0.8, "rationale": "..."}},
  "task_adherence": {{"score": 0.7, "rationale": "..."}},
  "task_completion": {{"score": 0.9, "rationale": "..."}}
}}
Do not include any extra text, markdown, or explanation outside the JSON object.
\end{lstlisting}

\section{Generative AI Usage Statement}

No Generative AI (GenAI) was used in the \textit{preparation} of this manuscript. GenAI was used though in a \textit{post-writing} step to perform an initial `review' and recommend a bullet list of improvements before the submission (Listing \ref{cod:review_prompt}).

\begin{lstlisting}[numbers={none}, caption={Prompt used to perform an initial 'review' before submission.}, label={cod:review_prompt}]
Create an academic paper review for the attached PDF file. 
Consider the criteria in the following CFP: {URL}. 
Present your feedback in a bullet list, reduced format.
\end{lstlisting}

\begin{landscape}
\small
\begin{table*}
  \caption{LLM-agnostic python agent frameworks (sorted by GitHub stars).}
  \label{tab:framewors}
  \begin{tabular}{lp{7cm}p{1.5cm}rrr}
    \toprule
    Name & Short description & License & \faGithub~Stars & \faGithub~Forks & \faGithub~Contributors \\ 
    \midrule
    AutoGPT \cite{Gravitas2025} & Framework for managing GPT-4 agents. & PolyForm 1.0 and MIT & 179k & 46k & 781 \\
    LangChain \cite{LangChain2025} & Framework for building applications through composability. & MIT & 133k & 21.9k & 3,914 \\
    AutoGen \cite{Wu2023} & Framework for creating MAS that can act autonomously or work alongside humans. & CC-BY-4.0 & 56.8k & 8.5k & 558 \\
    LlamaIndex \cite{Liu2022} & Data framework for LLM application allowinf for covering 
    core and plugin components. & MIT & 48.4k & 7.2k & 1,855 \\
    Mem0 \cite{Chhikara2025} & Memory layer for agents. & Apache 2.0 & 52.2k & 5.8k & 297 \\
    CrewAI \cite{Crewai2025} & Framework for orchestrating role-playing MAS. & MIT & 48.3k & 6.6k & 302 \\
    Agno \cite{Agno2025} & Framework for managing MAS in the cloud. & Apache 2.0 & 39.2k & 5.2k & 431 \\
    LangGraph \cite{LangGraph2025} & MAS orchestration framework for building, managing, and deploying long-running, stateful agents. & MIT & 28.6k & 4.9k & 289 \\
    Semantic Kernel \cite{SemanticKernel2025} & Orchestration framework for MAS & MIT & 27.7k & 4.5k & 435 \\
    Smolagents \cite{Smolagents2025} & Model-agnostic framework for MAS aiming for simplicity and multiple modalities. & Apache 2.0 & 26.5k & 2.4k & 208 \\
    Haystack \cite{Pietsch2019} & Orchestration framework to build customizable applications. & Apache 2.0 & 24.8k & 2.7k & 351 \\
    CAMEL \cite{Li2023} & MAS framework aiming at large-scale projects. & Apache 2.0 & 16.6k & 1.9k & 212 \\
    Pydantic AI \cite{PydanticAI2025} & Framework for agent workflows inspired by FastAPI development. & MIT & 16.2k & 1.9k & 425 \\ 
    Upsonic \cite{Upsonic2025} & Agent development framework aiming at safety-first agents. & MIT & 7.8k & <1k & 38 \\
    AutoAgent \cite{Tang2025} & Automated agent framework enabling creation and deployment of agents using natural language. & MIT & 9k & 1.3k & 5 \\
    Langroid \cite{Langroid2025} & A lightweight, principled framework for extensible creation of LLM-powered MAS. & MIT & 4k & <1k & 44 \\
    BeeAI \cite{IBMResearch2024} & Framework for building MAS. & Apache 2.0 & 3.2k & <1k & 70 \\
    AGiXT \cite{AGiXT2025} & Framework for orchestration of agents in virtual and physical environments. & MIT & 3.2k & <1k & 45 \\
  \bottomrule
\end{tabular}
\end{table*}
\end{landscape}

\begin{landscape}
\small
\begin{table*}
\caption{Benchmarks used in agent evaluations (alphabetical order).}
\label{tab:benchmarks}
\begin{tabular}{lp{7cm}p{4cm}l}
\toprule

Benchmark & Task & Environment & Modalities \\ 
\midrule

AgentBench \cite{Liu2023} & Multi-environment evaluation of LLMs acting as agents & Mixed (APIs, simulated env.) & Actions\\ 

AgentEval \cite{Arabzadeh2024} & Multi-dimensional utility assessment framework for LLM-powered applications & Offline simulation-based task evaluation & Text, Solution traces \\ 

Agent-SafetyBench \cite{Zhang2025} & Tasks to assess safety of agents & Simulated interaction env. & Text \\ 

AgentSims \cite{Lin2023} & Sandbox for creating agent evaluation tasks & Simulated GUI sandbox & Text, Actions \\ 

ALFWorld \cite{Shridhar2020} & Align text-based abstractions with embodied environments & Simulated & Text, Vision, Actions \\ 

ClemBench \cite{Chalamalasetti2023} & Game-based interactive tests for chat-optimized LLMs & Simulated games & Text \\ 

E2E \cite{Banerjee2023} & End-to-end benchmark and evaluation protocol for chatbots & Offline evaluation suite & Text \\ 

EmotionBench \cite{Huang2023} & Evaluate LLMs' emotional appraisal and empathy via scenarios & Eval framework & Text \\ 

Gentopia \cite{Xu2023} & Platform for tool-augmented LLM collaboration & Mixed (demos,  toolkit) & Text, Tool calls \\ 

IGLU \cite{Mehta2023} & Interactive collaborative environment for grounded language tasks & Collaborative env. & Text, Actions \\ 

ST-WebAgentBench \cite{Levy2025} & Tasks paired with safety and trustworthiness policies & Simulated interaction env. & Text \\

Mind2Web \cite{Deng2023} & Generalist web agent evaluation via real-world website tasks & Real web browsing & Text, Web actions \\ 

Mobile-Env \cite{Schneider2022} & Coordination environment for wireless mobile networks & Simulated wireless network env. & Actions \\ 

Tachikuma \cite{Liang2023} & Complex multi-character interactions benchmark for LLMs & Simulated interaction env. & Text \\ 

ToolBench \cite{Xu2023a} & API evaluation suite & API tasks & Text, API calls \\ 

PentestGPT \cite{Deng2023a} & Penetration-testing benchmark and system & Pentest targets & Text, Tool usage \\ 

RocoBench \cite{Zhao2023} & Multi-robot collaboration benchmark & Multi-robot tabletop simulation & Text, Actions \\ 

SafeArena \cite{Tur2025} & Assesses misuse of web agents towards safety & Simulated interaction & Text \\ 

SocKET \cite{Choi2023} & Wide-ranging social-language benchmark & Offline tasks & Text \\ 

WebArena \cite{Zhou2023} & Realistic web environment for building autonomous web agents & Real web browsing & Text, Web actions, Tools \\ 

WebShop \cite{Yao2023} & Large-scale simulated e-commerce environment for grounded agents & Simulated e-commerce site & Text, Web actions, Vision \\ 

\bottomrule
\end{tabular}

\end{table*}
\end{landscape}

\end{document}